\documentclass{article}

\usepackage[square,sort,comma,numbers]{natbib}
\usepackage[preprint]{neurips_2020}

\usepackage[utf8]{inputenc} 
\usepackage[T1]{fontenc}    
\usepackage{hyperref}       
\usepackage{url}            
\usepackage{booktabs}       
\usepackage{amsfonts}       
\usepackage{nicefrac}       
\usepackage{microtype}      

\usepackage{amsmath}
\usepackage{subcaption}
\usepackage[normalem]{ulem}
\usepackage{amssymb}
\usepackage{geometry}
\usepackage[colorinlistoftodos]{todonotes}

\usepackage{floatrow}
\def\mathclap#1{\text{\hbox to 0pt{\hss$\mathsurround=0pt#1$\hss}}}
\title{Set Distribution Networks: a Generative Model for Sets of Images}

\author{%
  Shuangfei Zhai, Walter Talbott, Miguel Angel Bautista, 
  Carlos Guestrin, Josh M. Susskind \\
  Apple Inc.\\
  \texttt{\{szhai,wtalbott,mbautistamartin,guestrin,jsusskind\}@apple.com} \\
}

\begin{document}

\maketitle

\begin{abstract}
Images with shared characteristics naturally form sets. For example, in a face verification benchmark, images of the same identity form sets. For generative models, the standard way of dealing with sets is to represent each as a one hot vector, and learn a conditional generative model $p(\mathbf{x}|\mathbf{y})$. This representation assumes that the number of sets is limited and known, such that the distribution over sets reduces to a simple multinomial distribution. In contrast, we study a more generic problem where the number of sets is large and unknown. 
We introduce Set Distribution Networks (SDNs), a novel framework that learns to autoencode and freely generate sets. We achieve this by jointly learning a set encoder, set discriminator, set generator, and set prior. We show that SDNs are able to reconstruct image sets that preserve salient attributes of the inputs in our benchmark datasets, and are also able to generate novel objects/identities. We examine the sets generated by SDN with a pre-trained 3D reconstruction network and a face verification network, respectively, as a novel way to evaluate the quality of generated sets of images.
\end{abstract}

\section{Introduction}
Generative modeling of natural images has seen great advances in recent years. State of the art models such as GANs \cite{biggan}, EBMs \cite{du2019implicit} and VAEs \cite{vqvae2} can generate single images with high perceptual quality. In many applications, however, images often come in sets with shared characteristics. For example, a set might be constructed from images that belong to the same semantic category, or those that share the same attribute. When the number of sets is limited and known, one is able to easily extend a generative model to its conditional version by representing the set information as a one hot vector \cite{cgan,cgan_proj}. We refer to these models as class conditional generative models (CCGMs). CCGMs are fundamentally limited by the pre-defined enumeration of all possible sets in their encoding, which prevents recognizing or generating sets that are not specifically encoded from the training distribution. Also, the number of parameters needed for a CCGM grows linearly w.r.t the number of sets (classes) during training, which limits its scalability.

In this paper, we study the problem of generative modeling of sets of images with a generic approach. We propose Set Distribution Networks (SDNs), a probabilistic model that is capable of learning to stochastically reconstruct a given set and generate novel sets at the same time.  Stochastic reconstruction of a set means that the individual images in the input set will not be reproduced exactly, but the generated images will share set-defining attributes with the input images.  We achieve this by jointly training a set encoder, a set discriminator, a set generator, and a set prior. We train an SDN by following the MLE objective, which results in an adversarial game where the encoder, discriminator and prior are trained against the generator.

We evaluate SDNs on two benchmarks: 1. ShapeNet \cite{shapenet}, which consists of objects in various viewpoints; 2. VGGFace2 \cite{vggface2}, which consists of various faces of human identities. We show that the same SDN architecture can be successfully trained on the two datasets, and can learn to both reconstruct an unseen set, and generate a novel set. We measure the quality of the set generative model by examining the generated samples with a pre-trained 3D reconstruction network, and face verification network, respectively, and show that SDNs learn to generate faithful and coherent sets.

\section{Methodology}
\subsection{Set Distribution Networks}
We denote an image set of size $n$ as $\mathbf{X} \in \mathcal{X}$, with $\mathbf{X} = \{\mathbf{x}_i\}_{i=1...n}$, $\mathbf{x}_i \in R^d$, and we are interested in learning a probabilistic model $p_{\theta}(\mathbf{X})$. To do so, we propose a novel encoder-decoder styled model, dubbed Set Distribution Networks (SDNs), that takes the form:
\begin{equation}\label{eq:sdn}
p_{\theta}(\mathbf{X}) = p_{\theta}(z(\mathbf{X};\theta))p_{\theta}(\mathbf{X}|z(\mathbf{X};\theta)).
\end{equation}
Here $z: \mathcal{X} \rightarrow \mathcal{Z}$ is a deterministic function that maps a set $\mathbf{X}$ to an element $\mathbf{z}$ in a discrete space $\mathcal{Z}$. $p_{\theta}(\cdot)$ is a prior distribution, with the support given by $supp(z) = \{z(\mathbf{X};\theta): \mathbf{X} \in \mathcal{X}\}$ which is a subset of $\mathcal{Z}$.  $p_{\theta}(\mathbf{X}|z(\mathbf{X}))$ is a conditional distribution which is defined as :
\begin{equation}\label{eq:decoder}
\begin{split}
p_{\theta}(\mathbf{X}|\mathbf{z}) &= \frac{I(z(\mathbf{X})=\mathbf{z})e^{-\sum_{\mathbf{x} \in \mathbf{X}}E(\mathbf{x}, \mathbf{z};\theta)}}{\int_{\mathbf{X'}\in \mathcal{X}}I(z(\mathbf{X'})=\mathbf{z})e^{-\sum_{\mathbf{x} \in \mathbf{X'}}E(\mathbf{x}, \mathbf{z}; \theta)} d\mathbf{X'}}.
\end{split}
\end{equation}
Here $I(\cdot)$ is the indicator function. The conditional probability takes the form of an energy based model (EBM), where density is only assigned to sets $\mathbf{X'}$ that are mapped to the same $\mathbf{z}$. 

We first show that Equation \ref{eq:sdn} indeed defines a valid distribution of $\mathbf{X}$. To see this, we  have:
\begin{equation}
\begin{split}
\int_{\mathbf{X} \in \mathcal{X}} p_{\theta}(\mathbf{X}) d \mathbf{X} &=  \int_{\mathbf{X} \in \mathcal{X}}  p_{\theta}(z(\mathbf{X};\theta))p_{\theta}(\mathbf{X}|z(\mathbf{X})) d \mathbf{X} = \sum_{\mathbf{z} \sim supp(z)} \int_{\mathbf{X} \in \mathcal{X}_{\mathbf{z}}}p_{\theta}(\mathbf{z})p_{\theta}(\mathbf{X}|\mathbf{z}) d\mathbf{X} \\
&= \sum_{\mathbf{z} \in supp(z)} p_{\theta}(\mathbf{z}) \int_{\mathbf{X} \in \mathcal{X}_{\mathbf{z}} }p_{\theta}(\mathbf{X}|\mathbf{z}) d\mathbf{X} =  \sum_{\mathbf{z} \in supp(z)} p_{\theta}(\mathbf{z})  = 1,
\end{split}
\end{equation}
where $\mathcal{X}_{\mathbf{z}}\triangleq \{\mathbf{X}:z(\mathbf{X};\theta)=\mathbf{z},\mathbf{X}\in \mathcal{X}\}$.
Here  we first partition the integration by grouping $\mathbf{X}$ that share the same $\mathbf{z}$, then move $p_{\theta}(\mathbf{z})$ out of the integral as it's a constant within the partition, and lastly recognize that the integration evaluates to 1 according to the definition of Equation \ref{eq:decoder}.

Intuitively, the SDN is an encoder-decoder model with discrete latent variables, with $z(\cdot;\theta)$ being the encoder and $p_{\theta}(\cdot|\cdot)$ being the probabilistic decoder.  The encoder serves the role of partitioning the input space, while the decoder defines a normalized distribution over sets mapped to the same partition. 

\subsection{Approximate Inference with Learned Prior and Generator}
We apply MLE to estimate the parameters of $p_{\theta}(\mathbf{X})$, where the negative log likelihood loss for an observed set $\mathbf{X}$ in the training split is:
\begin{equation}\label{eq:mle}
-\log p_{\theta}(z(\mathbf{X};\theta)) - \log p_{\theta}(\mathbf{X}|z(\mathbf{X};\theta)),
\end{equation}
which is decomposed into two parts, corresponding to the prior and decoder distribution respectively. 
We adopt a simple parameterization of the prior distribution as:
$
p_{\theta}(\mathbf{z}) = \frac{\bar{p}_{\theta}(\mathbf{z})}{\sum_{\mathbf{z} \in supp(z), }\bar{p}_{\theta}(\mathbf{z})}, \forall \mathbf{z} \in supp(z)$; $p_{\theta}(\mathbf{z}) = 0$ otherwise.
Here $\bar{p}_{\theta}(\mathbf{z})$ is a normalized distribution over $\mathcal{Z}$. Because $supp(z)$ is always a subset of $\mathcal{Z}$, we have that $p_{\theta}(\mathbf{z}) \ge \bar{p}_{\theta}(\mathbf{z}), \forall \mathbf{z} \in supp(z)$.
We then achieve an upper bound of -$\log p_{\theta}(z(\mathbf{X};\theta))$ given by 
\begin{equation}\label{eq:l0}
\mathcal{L}_0(\theta) \triangleq -\log \bar{p}_{\theta}(z(\mathbf{X};\theta)),
\end{equation}
which gets rid of the need for the intractable $supp(z)$. For -$\log p_{\theta}(\mathbf{X}|z(\mathbf{X};\theta))$, we apply variational inference with a learned generator, as in \cite{kimandbengio,ebgan,calibrating,afv}, where we have:
\begin{equation}\label{eq:l1}
\begin{split}
&-\log p_{\theta}(\mathbf{X}|z(\mathbf{X};\theta)) = -\log\frac{I(z(\mathbf{X};\theta)=\mathbf{z})e^{-\sum_{\mathbf{x} \in \mathbf{X}}E(\mathbf{x}, \mathbf{z};\theta)}}{\int_{\mathbf{X'}\in \mathcal{X}}I(z(\mathbf{X'};\theta)=\mathbf{z})e^{-\sum_{\mathbf{x} \in \mathbf{X'}}E(\mathbf{x}, \mathbf{z};\theta)} d\mathbf{X'}} \Bigg |_{\mathbf{z}=z(\mathbf{X};\theta)}\\
&= \sum_{\mathbf{x} \in \mathbf{X}}E(\mathbf{x}, \mathbf{z};\theta) + \log \int_{\mathbf{X'}\in \mathcal{X}}I(z(\mathbf{X'};\theta)=\mathbf{z})e^{-\sum_{\mathbf{x} \in \mathbf{X'}}E(\mathbf{x}, \mathbf{z};\theta)} d\mathbf{X'} \Bigg |_{\mathbf{z}=z(\mathbf{X};\theta)}\\
&\geq  \sum_{\mathbf{x} \in \mathbf{X}}E(\mathbf{x}, \mathbf{z};\theta) +\mathrm{E}_{\mathbf{X'}\sim p_{\psi}(\mathbf{X}|\mathbf{z})}\log [I(z(\mathbf{X'};\theta)=\mathbf{z})e^{-\sum_{\mathbf{x} \in \mathbf{X'}}E(\mathbf{x}, \mathbf{z};\theta)}] + H(p_{\psi}) \Bigg |_{\mathbf{z}=z(\mathbf{X};\theta)}\\
&\triangleq \mathcal{L}_1(\theta, \psi)
\end{split}
\end{equation}Here we have derived a lower bound of -$\log p_{\theta}(\mathbf{X}|\mathbf{z})$\footnote{we use $\mathbf{z}$ in place of $z(\mathbf{X};\theta)$ when there is no ambiguity} by introducing a variational distribution $p_{\psi}(\mathbf{X}|\mathbf{z})$, which we parameterize in the form of a generator:
\begin{equation}\label{eq:g}
\int_{\mathbf{X} \sim p_{\psi}(\mathbf{X}|\mathbf{z})}f(\mathbf{X}) d\mathbf{X} \triangleq \int_{\{\mathbf{z'}_i \sim p_{\psi}(\mathbf{z'})\}_{i=1...n}} f(\{G(\mathbf{z,z'_i};\psi)\}_{i=1...n})d\{\mathbf{z'}_i \}_{i=1...n}, \forall f.
\end{equation}Here $p_{\psi}(\mathbf{z'})$ is a simple distribution (e.g., Isotropic Gaussian) over $\mathcal{Z'}$, and $G: \mathcal{Z}\times \mathcal{Z'} \rightarrow R^d$ is a deterministic function (the generator) that maps a $(\mathbf{z}, \mathbf{z'})$ tuple to the image space.  The lower bound $\mathcal{L}_1(\theta, \psi)$ can be tightened by solving $\max_{\psi}\mathcal{L}_1(\theta, \psi)$ \cite{afv}.

\subsection{Model Architectures}
\begin{figure}
\centering
\includegraphics[scale=0.2]{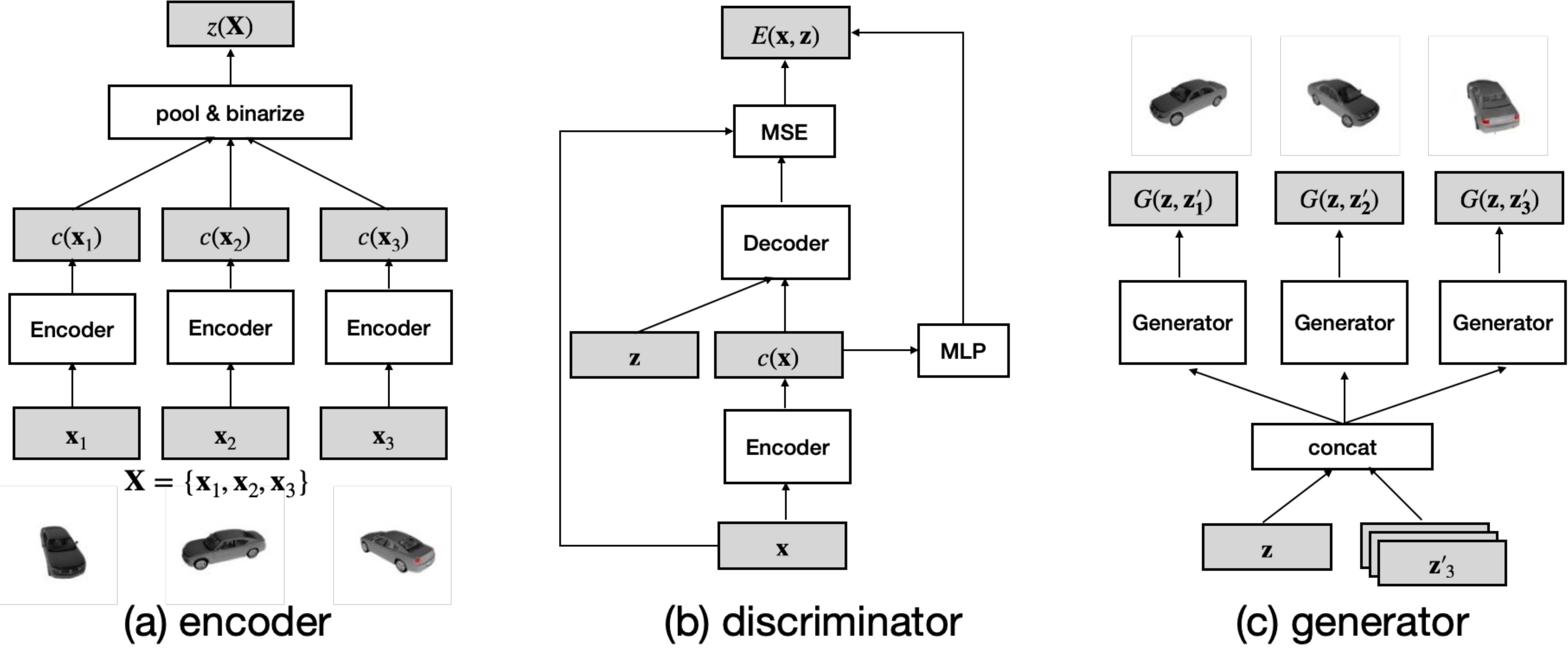}
\caption{Model architectures of SDNs. SDNs consist of three modules: (a) a set encoder that maps a set of images into a discrete code, with a shared convolutional encoder encoding each image followed by average pooling and discretization; (b) a conditional discriminator (energy) for each image that  takes the form of an autoencoder (similar to \cite{ebgan}); (c) a conditional generator that generates images conditioned on a set code $\mathbf{z}$.}
\label{fig:arch}
\end{figure}

There are four interacting modules for SDNs: the prior $p_{\theta}(\mathbf{z})$, the encoder $z(\mathbf{X};\theta)$, the discriminator (energy) $E(\mathbf{x, z};\theta)$ and the generator $G(\mathbf{z, z'};\psi)$. We now explain the architecture and design choice for each of them. See Figure \ref{fig:arch} for an illustration.

\textbf{Prior}. In all of our implementations, we adopt binary set codes, i.e., letting $\mathcal{Z} = \{-1, 1\}^{d_z}$. In theory, one can choose any prior distribution over discrete variables. We use a standard auto-regressive model MADE \cite{made} with three fully connected layers, mainly for its simplicity and robustness.

\textbf{Encoder}. As a necessary condition, an encoder for a set needs to satisfy the permutation invariant property \cite{deepset}. We opt to use a simple architecture design where we let $z(\mathbf{X};\theta) = binarize(\frac{1}{n}\sum_{i=1}^n c(\mathbf{x}_i; \theta))$, where $c(\cdot;\theta)$ is a standard CNN image encoder. The outputs of the image codes are averaged across the set and then passed to a differentiable binarization operator (with straight through gradient estimation) to produce the final set code. 

\textbf{Discriminator}. The discriminator's job is to assign low energy to observed images and high energy to generated images, given a set code $\mathbf{z}$. We use an autoencoder based energy function implementation, similar to \cite{ebgan}. We have found that this choice is important as it enables effective learning in early stages of training. We reuse $c(\cdot;\theta)$ as the encoder, and separately learn a decoder $d(\mathbf{z},c(\mathbf{x}))$, which takes the concatenation of the binary set code $\mathbf{z}$ and dense image code $c(\mathbf{x})$ as input, and outputs a ``reconstruction" of $\mathbf{x}$. We additionally learn a unary energy term of $\mathbf{x}$ with a small MLP $d_0: R^{d_z} \rightarrow R$ that takes $c(\mathbf{x})$ as input and outputs a scalar. This gives us the final output of the discriminator as:
\begin{equation}\label{eq:energy}
E(\mathbf{x, z};\theta) = \|\mathbf{x} - d(\mathbf{z}, c(\mathbf{x};\theta); \theta)\|_2^2 + d_0(c(\mathbf{x};\theta);\theta).
\end{equation}\textbf{Generator}. The generator generates a set conditioned on a set code $\mathbf{z}$ by sampling $n$ random variables $\{\mathbf{z'}_i \sim p_{\psi}(\mathbf{z}')\}_{i=1...n}$, each of which is concatenated with $\mathbf{z}$ and generates an image independently. 

\subsection{Losses}
SDNs consist of two sets of parameters to be optimized, $\theta$ and $\psi$, where $\theta$ denotes the combined parameters for the prior, encoder and discriminator; and $\psi$ denotes those for the generator. During training, with $\theta$ fixed, we first optimize $\psi$ to tighten the lower bound by solving $\max_{\psi}\mathcal{L}_1(\theta, \psi)$. In order to make this practical, we make two simplifications. First, we do not explicitly include the entropy term $H(p_{\psi})$. This seems problematic at first glance as $H(p_{\psi})$ plays the role of encouraging $p_{\psi}(\mathbf{X}|\mathbf{z})$ to have high entropy which prevents the mode collapse problem. However, we have found that we can implicitly achieve high diversity of the generated samples with careful learning scheduling and architectural design choices as commonly used in GAN training, see Sec. \ref{sec:implementation} for more details. Also refer to \cite{afv} for more justification of this choice. Second, the indicator function $I(z(\mathbf{X}') = \mathbf{z})$ which ensures that the generated sets are mapped to the same set code $\mathbf{z}$ is not differentiable. We thus choose to use a soft approximation: $I(z(\mathbf{X}') = \mathbf{z}) \approx e^{- \|[z(\mathbf{X'};\theta)\odot \mathbf{z}]_- \|_1}$, where $\odot$ is the element-wise product, $[\cdot]_{-}$ is the operator that zeros out the positive elements, and $\|\cdot\|_1$ is the $\ell_1$ norm. Intuitively, this approximation equates the indicator function when $sign(z(\mathbf{X'};\theta)) = \mathbf{z}$, and induces a value in $(0, 1)$ otherwise. This leads us to the final form of the loss for the generator:
\begin{equation}\label{eq:loss_g}
\begin{split}
\mathcal{L}_{\psi} &=-\mathrm{E}_{\{\mathbf{z'}_i \sim p_{\psi}(\mathbf{z'})\}_{i=1...n}}\log [e^{-\|[z(\mathbf{X'};\theta)\odot \mathbf{z}]_-\|_1}e^{-\sum_{\mathbf{x} \in \mathbf{X'}}E(\mathbf{x}, \mathbf{z};\theta)}]|_{\mathbf{X'} = \{G(\mathbf{z},\mathbf{z'}_i;\psi)\}_{i=1...n}}\\
&= \mathrm{E}_{\{\mathbf{z'}_i \sim p_{\psi}(\mathbf{z'})\}_{i=1...n}} [\|[z(\mathbf{X'};\theta)\odot \mathbf{z}]_-\|_1 + \sum_{\mathbf{x} \in \mathbf{X'}}E(\mathbf{x}, \mathbf{z};\theta)]|_{\mathbf{X'} = \{G(\mathbf{z},\mathbf{z'}_i;\psi)\}_{i=1...n}}.
\end{split}
\end{equation}
With $\psi$ fixed, we can optimize $\theta$ with a loss that combines $\mathcal{L}_0(\theta)$ and $\mathcal{L}_1(\theta,\psi)$. As is common practice in GAN and deep EBM training \cite{biggan,du2019implicit,ebgan}, we apply a margin based loss on $E(\mathbf{x},\mathbf{z};\theta)$. In particular, we have found that it's beneficial to apply two separate margins on the two terms of $E$, this leads to the loss function of $\theta$ update as:
\begin{equation}\label{eq:loss_d}
\begin{split}
\mathcal{L}_{\theta} = &-\log \bar{p}_{\theta}(\Omega(z(\mathbf{X};\theta))) \\
&+ \sum_{\mathbf{x} \in \mathbf{X}} \|\mathbf{x} - d(z(\mathbf{X};\theta), c(\mathbf{x};\theta); \theta)\|_2^2 + [\gamma_0 + d_0(\mathbf{x};\theta)]_{+} \\
&+\mathrm{E}_{\mathbf{X'} \sim p_{\psi}(\mathbf{X}|z(\mathbf{X}))} \sum_{\mathbf{x} \in \mathbf{X'}}[\gamma_1 - \|\mathbf{x} - d(z(\mathbf{X};\theta), c(\mathbf{x};\theta); \theta)\|_2^2]_+ +  [\gamma_0 - d_0(\mathbf{x};\theta)]_{+}.
\end{split}
\end{equation}
Here $\Omega$ is the stop gradient operator, indicating that we do not pass gradient from the prior to the encoder. $[\cdot]_+$ is the operator that zeros out the negative portions of the input. $\gamma_0, \gamma_1 \in R^+$are two margin parameters. 

\section{Related Work}
\textbf{Generative Modeling of sets}.
Generative modeling of sets is a challenging task. Hierarchical Bayes models have previously been widely applied to document modeling, which can be considered as a set of words, represented by the LDA model \cite{lda}. Recently, there has been a a body of work dedicated to point cloud generative modeling, where sets are composed of 3D points in Euclidean space \cite{pcgan,pcae,pointflow}. However, image sets considered in our work contain much richer structure than 3D points. It is thus unclear how methods developed from the point cloud or document modeling lines of work generalize to more complex image data.

\textbf{Autoencoding GANs}.
SDNs are similar to a handful of existing works which involve an encoder, generator/decoder and a discriminator \cite{bigan,vaegan,bigbigan}. Notably, BiGAN \cite{bigan} trains a two channel discriminator that tries to separate the two joint distributions $p(\mathbf{z})p(\mathbf{x}|\mathbf{z})$ and $p(\mathbf{x})p(\mathbf{z}|\mathbf{x})$. Aside from the set modeling aspect, in a BiGAN the encoder and decoder work jointly against the discriminator, whereas in SDNs, the encoder and the discriminator work jointly against the generator. Intuitively, SDNs encourage the encoder to learn discriminative representations by not allowing it to cooperate with the generator, which in turn improves the generator's quality as well.

\textbf{Conditional GANs}.
A conditional GAN (cGAN) \cite{cgan,cgan_proj} modifies the generator and discriminator of a GAN to be conditional on class labels. SDNs can be considered a generalization of cGANs, as it learns the parametric set representation jointly with the generator and discriminator. 

\textbf{VQ-VAEs}.
VQ-VAEs \cite{vqvae,vqvae2} are related to SDNs in that they learn encoders that output discrete and deterministic codes. They also learn auto-regressive priors which greatly improves their capability of generative modeling. SDNs differ from VQ-VAEs in that they have a more sophisticated encoder and discriminator, which are essential to cope with set structures.

\textbf{}

\section{Experiments}
\subsection{Datasets}
The first dataset we use is ShapeNet \cite{shapenet}, which consists of 3D objects with projected 2d views. We use a subset which contains 13 popular categories, namely \textit{airplane, bench, cabinet, car, chair, monitor, lamp, speaker, gun, couch, table, phone, ship}. The training set consists of 32,837 unique objects and 788,088 images in total. The second dataset is VGGFace2 \cite{vggface2}, which is a face dataset containing 9,131 subjects and 3.31M images.

For both datasets, we randomly construct fixed-size sets of 8 elements for training. For ShapeNet, this is done by randomly selecting an object and 8 of its views; for VGGFace2, we similarly select one identity and 8 of her/his faces. A mini-batch is constructed by selecting $N$ objects/subjects, which amounts to $N\times8$ images in total. All images are scaled to a size of $128\times128$.

\subsection{Implementation Details}
\label{sec:implementation}
Our implementation resembles that of SAGAN \cite{sagan} w.r.t. base architecture and learning scheduling. The encoder is identical to the discriminator in SAGAN, namely a CNN enhanced with Spectral Normalization (SN) \cite{sn} and Self Attention (SA), except that the number of output units is changed to $d_z$ instead of 1. The dimension of $\mathbf{z}$ is 2048 and 256 for ShapeNet and VGGFace2, respectively. The generator is also the same as that in SAGAN, which is a CNN with SN, Batch Norm \cite{bn} and SA. The decoder for the discriminator resembles the generator, but without BN and SA; the MLP for the unary energy term is a simple two layer ReLU neural network with $d_z$ hidden units. All images are normalized to range $[-1, 1]$. We set the two margin parameters as $\gamma_0 = 1, \gamma_1 = 0.1$, which we have found to work robustly across various settings.

During training, we use Adam with a learning rate 1e-4 for the generator , and learning rate 4e-4 for the encoder, discriminator, and prior. The momentum terms are set as $\{0, 0.999\}$ for both optimizers. We use a batch size of 32 (consisting of 256 images), spread across 4 V100 GPUs. We alternatively update $\psi$ and $\theta$ one step per mini-batch, and train the models for approximately 600K iterations, which takes roughly two weeks. 

\subsection{Qualitative Results}
\begin{figure}
\includegraphics[scale=0.38]{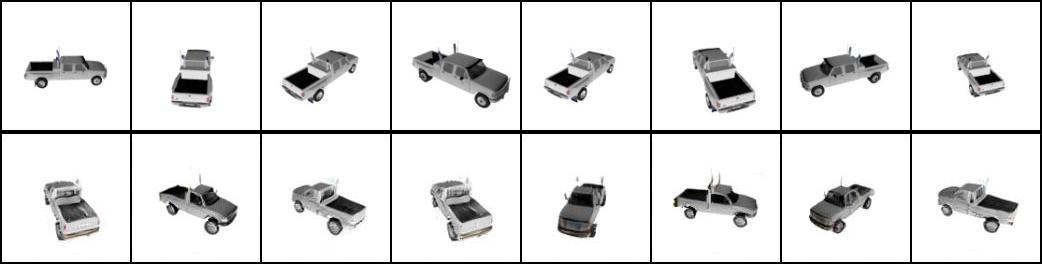}
\caption{Top: example set from the test split of ShapeNet; bottom: reconstructed set. There is no correspondence between images from the two sets. }
\label{fig:shapenet_recon}
\end{figure}

\begin{figure}
\floatbox[{\capbeside\thisfloatsetup{capbesideposition={right,top},capbesidewidth=4cm}}]{figure}[\FBwidth]
{\caption{(a) Input, and (b) reconstructed sets from a ShapeNet car where samples are arranged on a circle by manually estimated pose (see Supp. Material for details.) This visualization shows consistency in terms of appearance and pose variability between input and reconstructed sets.}\label{fig:car_wheel}}
{\includegraphics[width=8cm]{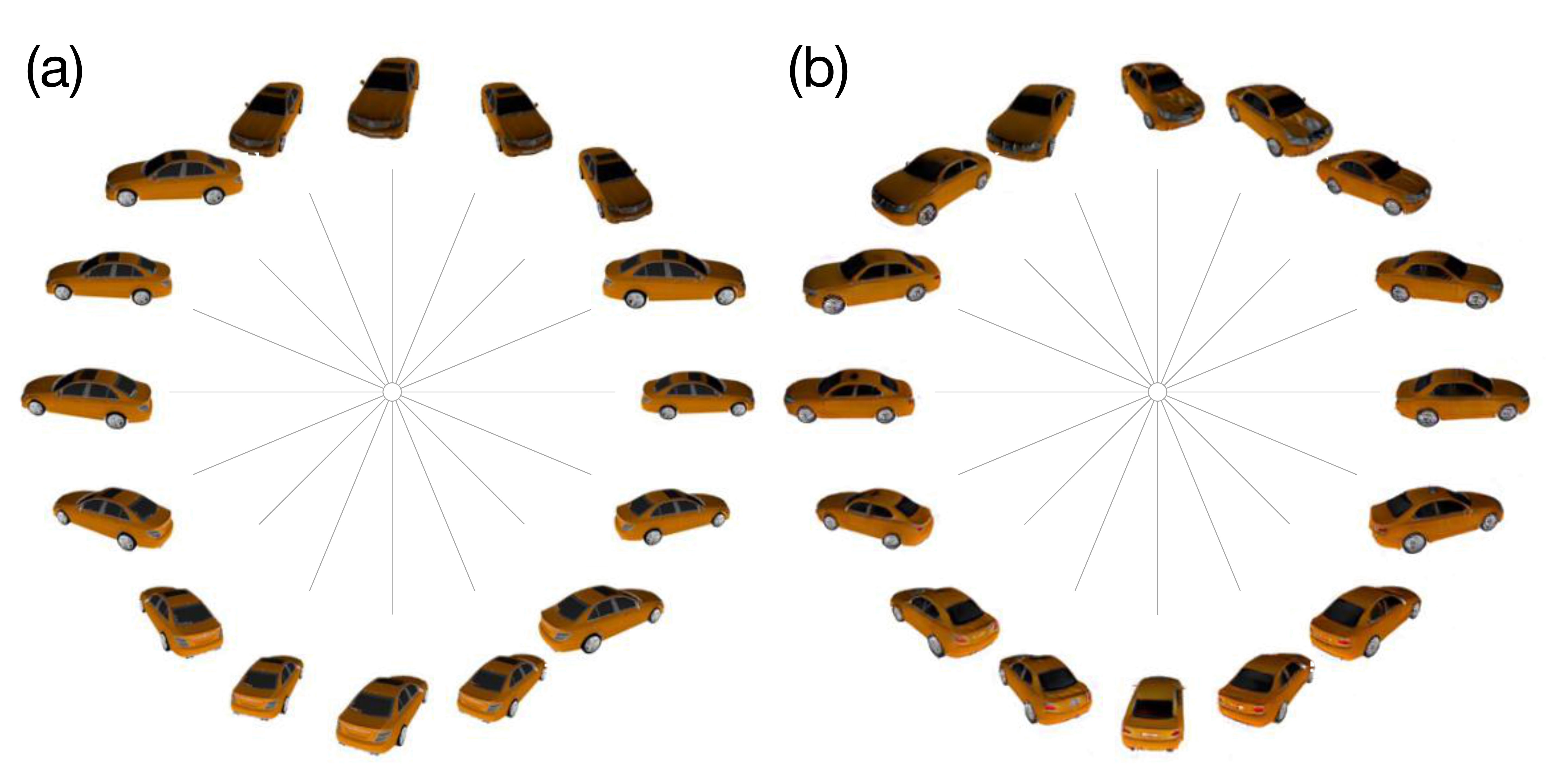}}
\end{figure}

\begin{figure}
\includegraphics[scale=0.38]{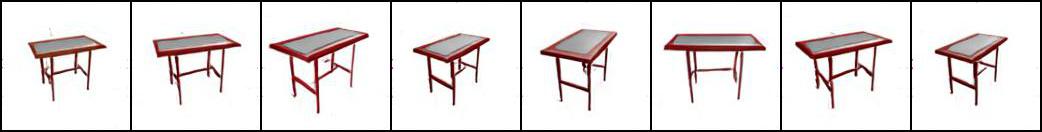}
\includegraphics[scale=0.38]{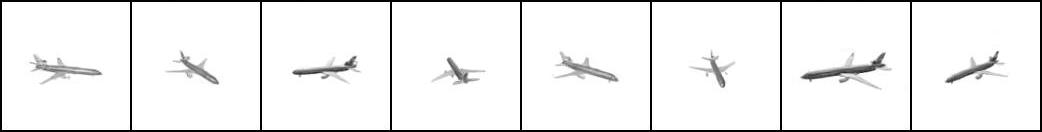}
\caption{Two sets sampled from the prior on ShapeNet.}
\label{fig:shapenet_gen}
\end{figure}

\begin{figure}
\includegraphics[scale=0.38]{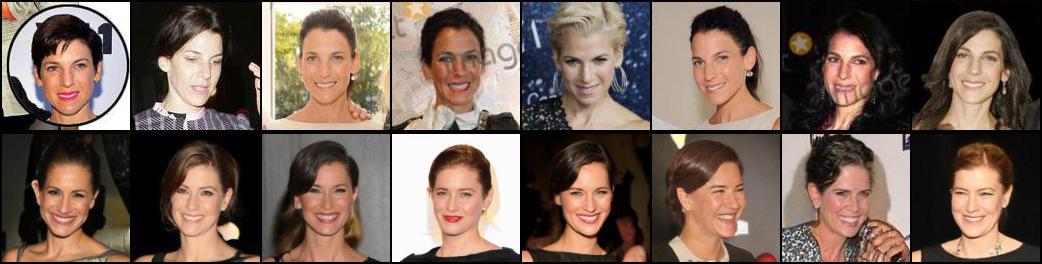}
\caption{Top: an example set from the test split of VggFace2; bottom: the reconstructed set. There is no correspondence between images from the two sets. }
\label{fig:vggface_recon}
\end{figure}

\begin{figure}
\includegraphics[scale=0.38]{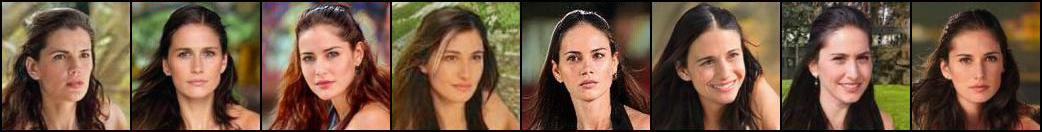}
\includegraphics[scale=0.38]{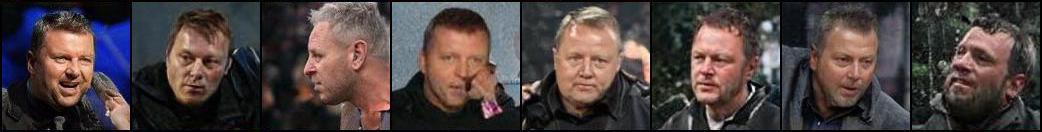}
\caption{Two sets sampled from the prior on VGGFace2.}
\label{fig:vggface_gen}
\end{figure}

By construction, SDNs are able to reconstruct (stochastically) an input set, and generate new sets by sampling from the prior. We show example inputs, reconstructions and samples in Fig. \ref{fig:shapenet_recon}, \ref{fig:shapenet_gen}, \ref{fig:vggface_recon}, \ref{fig:vggface_gen}. We see that SDNs produce semantically meaningful, consistent reconstructions of unseen input sets. The samples from the prior are also coherent. For faces, the SDN captures variations in hairstyle, expression, and pose, while retaining gender and ethnicity of the input identities. For ShapeNet, we show in Fig. \ref{fig:car_wheel} that the reconstructed sets match their input counterparts in terms of object consistency and pose diversity. See Supp. Materials for more qualitative results. 

\subsection{ShapeNet Evaluations}

In order to show that ShapeNet sets reconstructed by SDNs are consistent quantitatively, we take a pre-trained Occupancy Network \cite{occnet} (OccNet) and use it to generate 3D meshes from images of both real and SDN-reconstructed sets, where SDNs and OccNets are trained on the same train/test splits and rendered views. We extend OccNets \cite{occnet} to deal with multiple views by a simple average pooling of its encoder output across views, and refer readers to \cite{occnet} and the Supp. Material for more details.

We empirically evaluate SDNs by computing Chamfer Distance\footnote{We report CD$\times 10^3$} (CD) \cite{pcae} and IoU as in \cite{occnet} between corresponding 3D meshes obtained from real and SDN-reconstructed sets. We denote these meshes as $\mathcal{M}$ and $\hat{\mathcal{M}}$, respectively. We compute the CD on $2048$ points sampled from the surface of each mesh. To approximate the IoU we sample $2048$ points on the volume of each mesh and compute the average proportion of points of $\mathcal{M}$ contained in the volume of $\hat{\mathcal{M}}$ and vice-versa. We expect that if SDN-reconstructed sets represent accurate and consistent views of the same object, both metrics will improve as the difference between reconstructed and real sets decreases. 

We verify this hypothesis with two experiments, both conducted on the test split of ShapeNet. In the first experiment we fix the size of real sets to $16$ views and vary the number of views sampled from the reconstructed set\footnote{Note that the model is trained with sets of 8 but naturally generalizes to different set sizes at inference time, due to the use of average pooling in the encoder.}. We then use the sets to compute meshes $\mathcal{M}$ and $\hat{\mathcal{M}}$. We see that as the number of sampled reconstructed views increases, we consistently get better reconstructions, as shown in Tab. \ref{tab:shapenet_exp_1} (a). 

\begin{table}[h]
    \small
    \centering
    \begin{tabular}{cc}
    \begin{tabular}{c|c|c|c|c}
         \# Recon. views & 1 & 4 & 8 & 16  \\ \hline
         CD $\downarrow$ & 4.458  & 3.682 & 3.501 & 3.433\\ \hline
         IoU $\uparrow$ & 0.679  & 0.711  & 0.719  & 0.722 \\ 
    \end{tabular}
    & 
    \begin{tabular}{c|c|c|c|c}
         \# Input views & 1 & 4 & 8 & 16  \\ \hline
         CD $\downarrow$ & 6.333 & 3.912  & 3.590 & 3.434 \\ \hline
         IoU $\uparrow$ & 0.633  & 0.701 & 0.715 & 0.722
    \end{tabular}\\
    (a) & (b)
    \end{tabular}

    \caption{(a) 3D reconstruction results as a function of the reconstructed set size.  (b) 3D reconstruction results as a function of the number of samples $n$ used in the compute the set code $z(\textbf{X})$.}
    \label{tab:shapenet_exp_1}
\end{table}

\begin{figure}
    \centering
    \includegraphics[width=\textwidth]{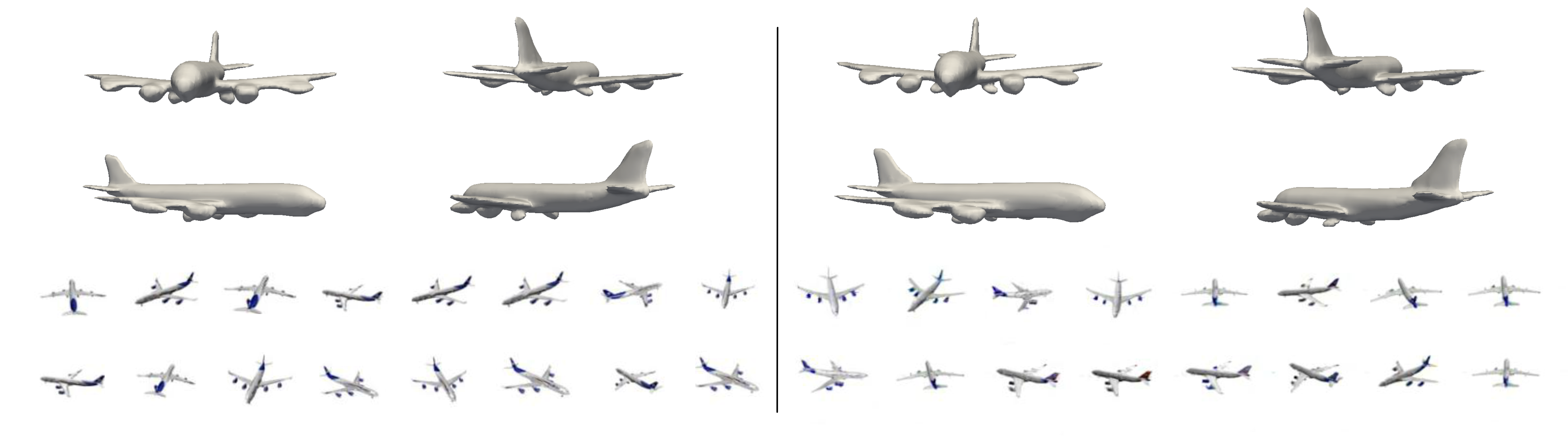}
    \caption{Left: real set and 3D reconstruction. Right: reconstructed set and 3D reconstruction. For this example the IoU and CD are $0.784$ and $4.289$, respectively.}
    \label{fig:3d_reconstructions}
\end{figure}

For our second experiment we vary the number of input views used to compute the set code $z(\textbf{X})$, while keeping the number of reconstructed views constant (16). We compute the reconstructed meshes and compare them with the corresponding mesh obtained with all 16 input views. Our hypothesis is that the set representation gets better as the number of samples in $\textbf{X}$ increases, and therefore sets can be more accurately reconstructed, which will lead to better 3D reconstructions, as shown in Tab. \ref{tab:shapenet_exp_1}(b). Finally, in Fig. \ref{fig:3d_reconstructions} we show qualitative comparisons of 3D meshes obtained from real vs. reconstructed sets, where both contain 16 elements.

\subsection{VGGFace2 Evaluations}
To provide quantitative results of set reconstruction performance for faces, we examine identity verification ROC curves on the VGGFace2 dataset using a pre-trained face verification model from \cite{vggface2}.  We evaluate both reconstructed samples and samples generated from the prior, corresponding to Fig. \ref{fig:vggface_recon} and Fig. \ref{fig:vggface_gen}, respectively. All input sets are taken from the test split of VGGFace2.

\begin{figure}
\begin{subfigure}{0.325\textwidth}
\includegraphics[scale=0.245]{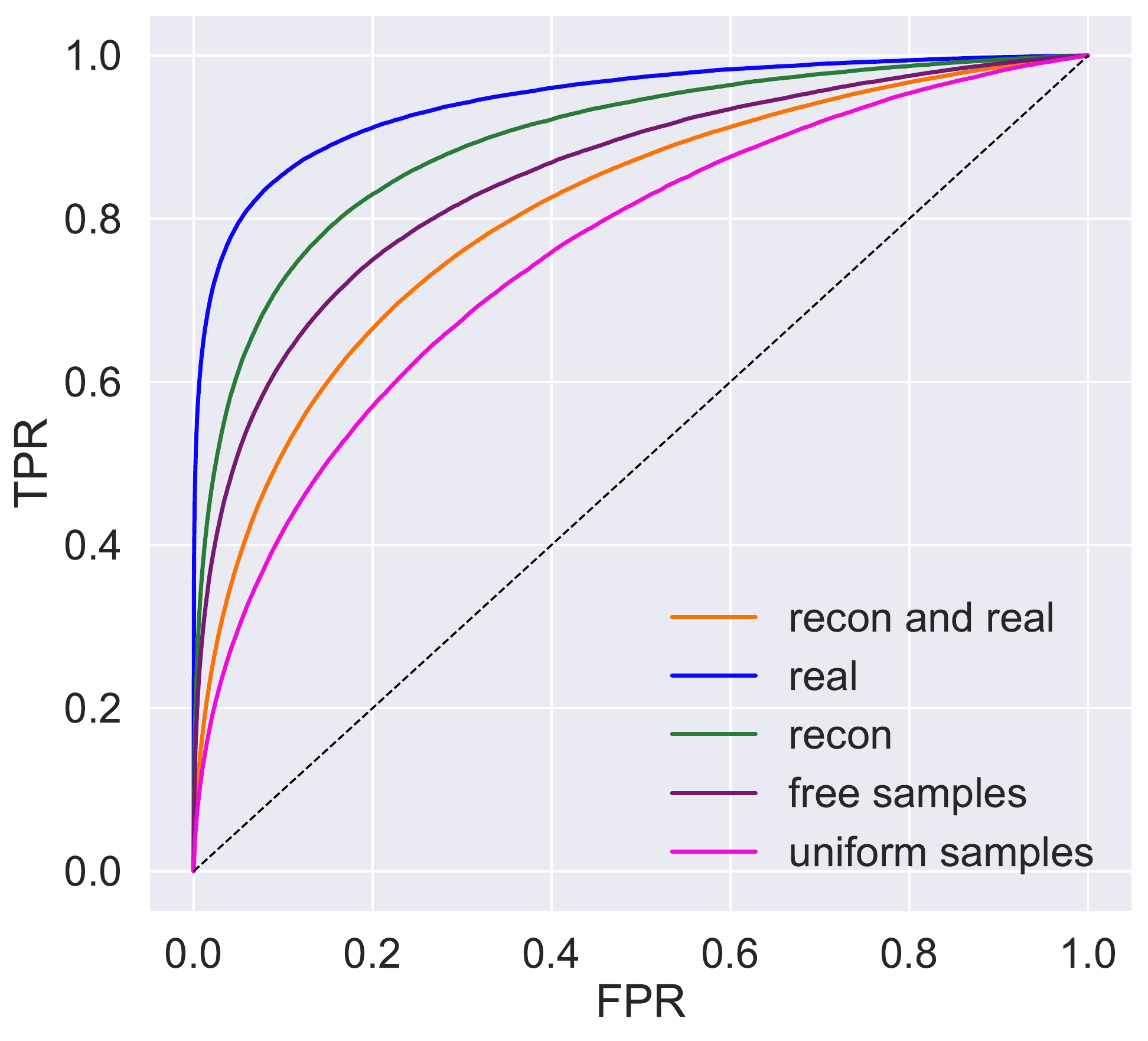}
\caption{Image type}
\label{fig:roc_sample_type}
\end{subfigure}
\begin{subfigure}{0.325\textwidth}
\includegraphics[scale=0.245]{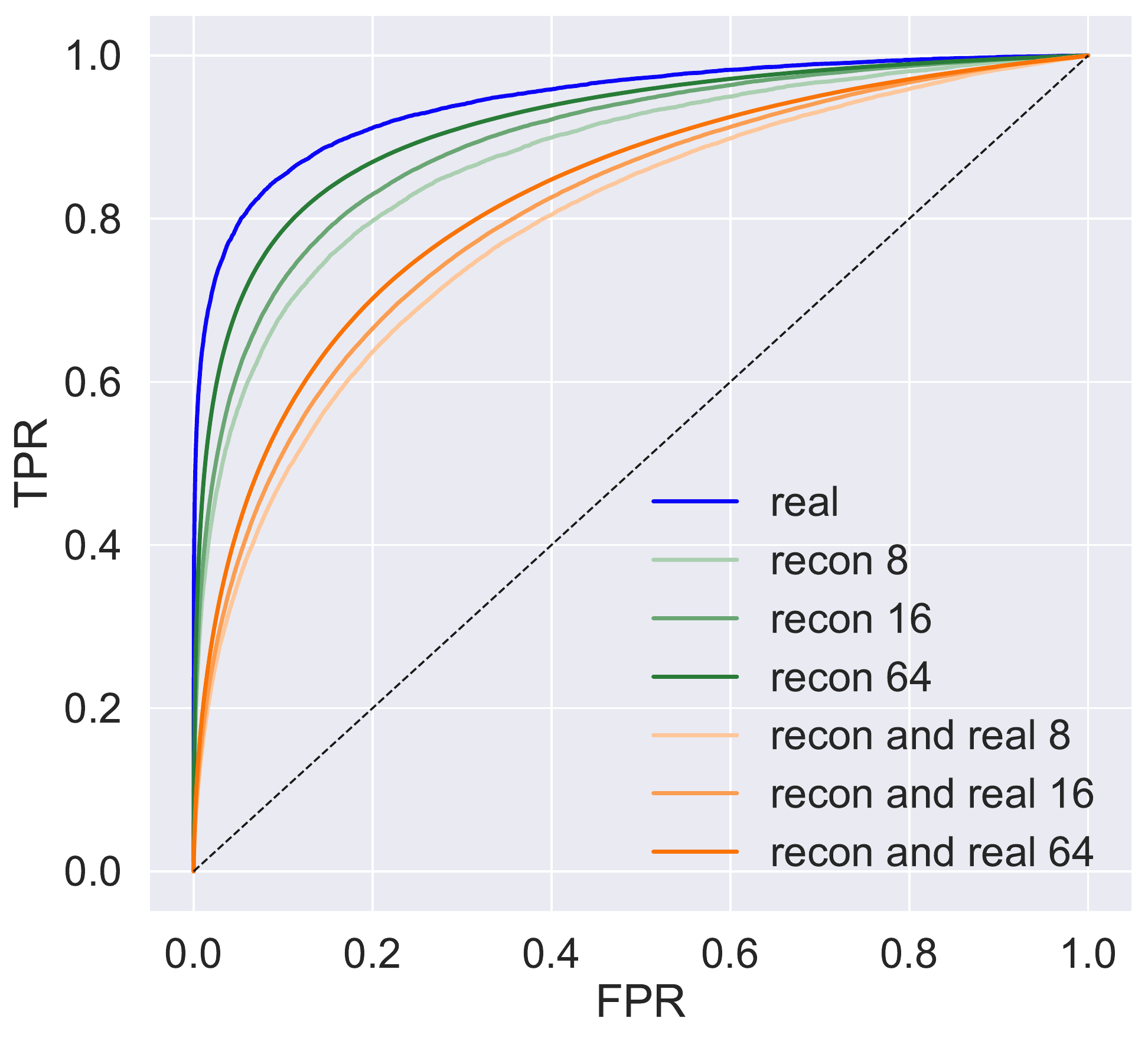}
\caption{Context size}
\label{fig:roc_context_size}
\end{subfigure}
\begin{subfigure}{0.325\textwidth}
\includegraphics[scale=0.245]{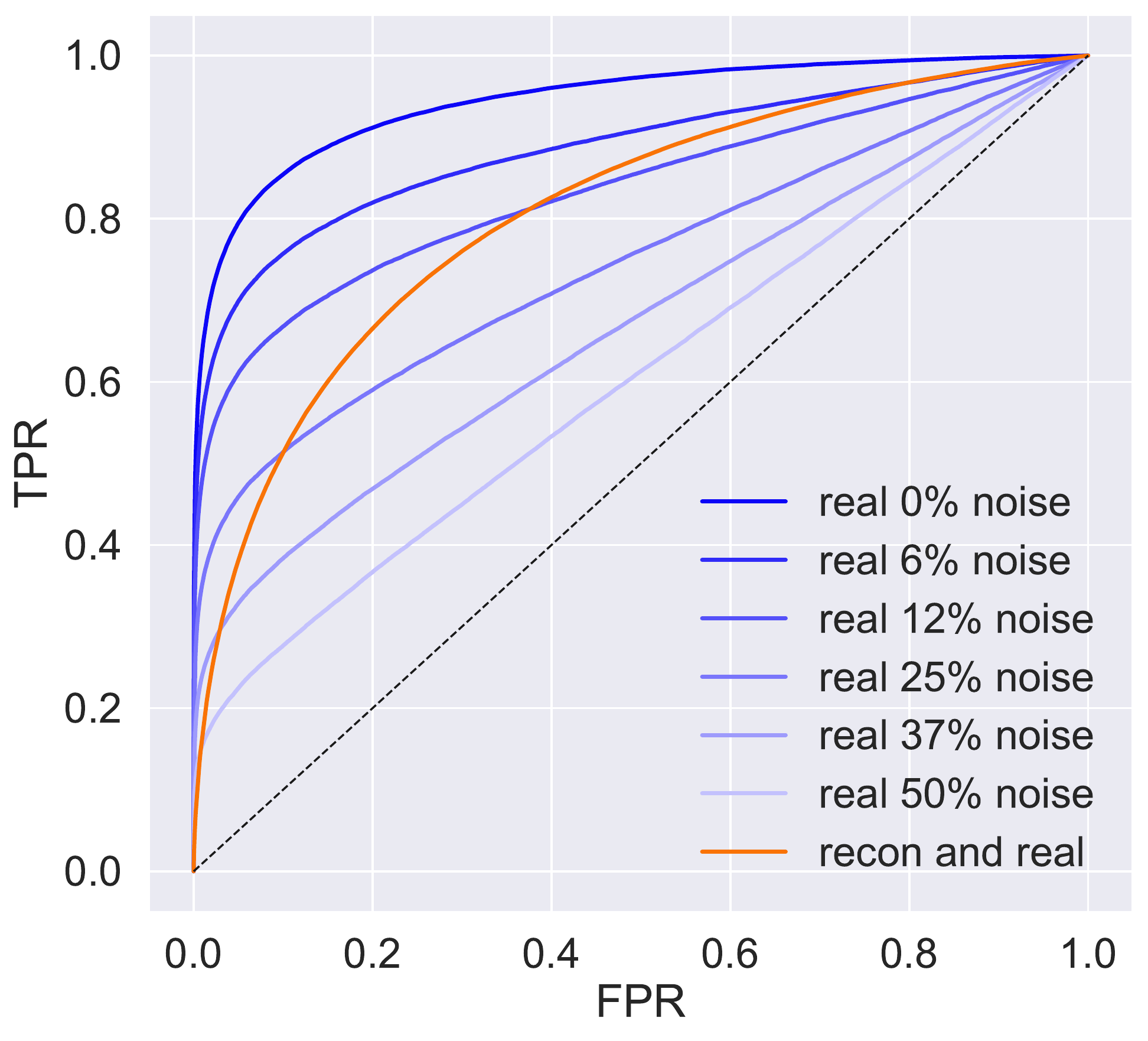}
\caption{Label noise}
\label{fig:roc_contamination}
\end{subfigure}
\caption{Identity verification performance with a pre-trained VGGFace2 model. 'recon and real': matching reconstructed images to real images; 'real': matching real images; 'recon': matching reconstructed images; 'free samples': matching samples generated from the learned prior. 'uniform samples':  matching samples generated from a uniform prior.}
\end{figure}

As a preprocessing step, we convert each image (real and generated) into a fixed dimensional embedding with the VGGFace2 model. We then use the distances in the embedding space to compare same and different identities' true-positive rate (TPR) and false-positive rate (FPR).  We plot the ROC curve for different types of input images in Figure~\ref{fig:roc_sample_type}. We see that the curve for reconstructed images ('recon') is close to that of real images ('real'), suggesting that the reconstructed sets are diverse and also that the images within a set are consistent with each other.  Comparing the reconstruction to real images ('recon and real'), we see that the reconstructions are diverse and self-consistent, although the SDN does not strictly preserve identity in the reconstructed sets. As a better calibration of this difference, we also show the ROC for real images with different proportions of label noise in Figure~\ref{fig:roc_contamination}. Here we randomly contaminate the sets of real images with increasing numbers of images from different identities.  We can see that the reconstruction performance at different FPR thresholds approximates different levels of label noise.  For example, at a 10\% FPR, we see that the reconstruction sets perform similarly to real identities with 25\% label noise. 

Additionally, we look at the effect of context size on the reconstructed images in Figure~\ref{fig:roc_context_size}.  Even though the SDN was trained with a constant context size of 8, increasing the size of the input set improves the performance, and here we see reconstructed images approach even closer the performance of the model on the real images, which is also consistent with Table \ref{tab:shapenet_exp_1}.

\section{Conclusion}
In this paper, we presented SDNs, a novel probabilistic generative model for image sets. We demonstrated that SDNs can be successfully trained on two real world datasets, ShapeNet and VGGFace2, at $128\times 128$ resolution. We proposed to evaluate trained SDNs with a pretrained 3D reconstruction network and a face verification network, respectively, and showed that the trained SDNs generate high quality image sets both qualitatively and quantitatively.

\medskip
\small
\bibliographystyle{plain}
\bibliography{refs}
\appendix
\section*{Appendix}
\section{More samples}
\begin{figure}
    \centering
    \includegraphics[scale=0.28]{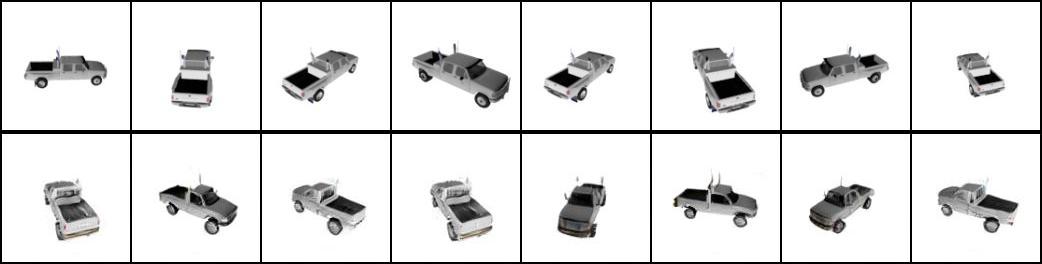}
        \includegraphics[scale=0.28]{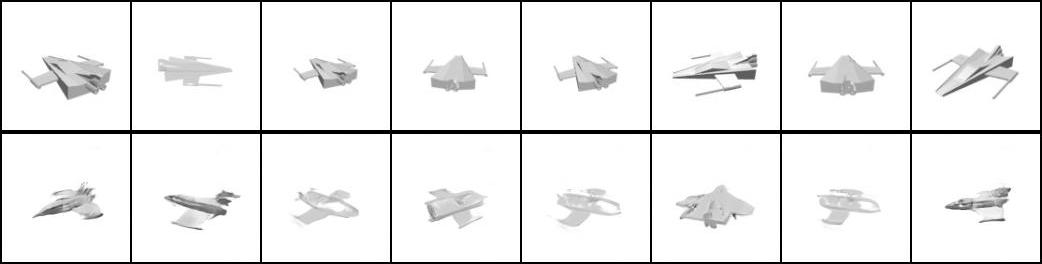}    \includegraphics[scale=0.28]{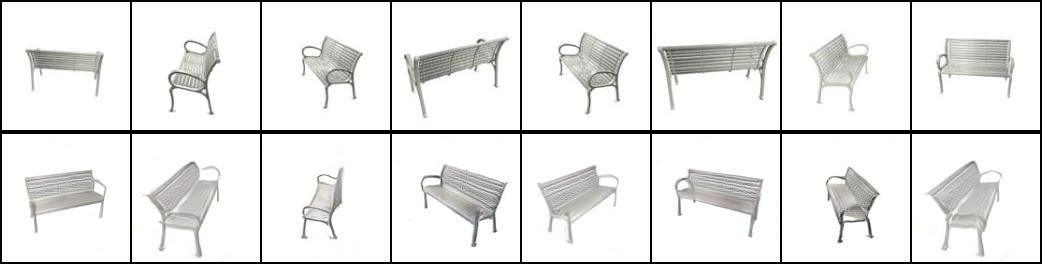}    \includegraphics[scale=0.28]{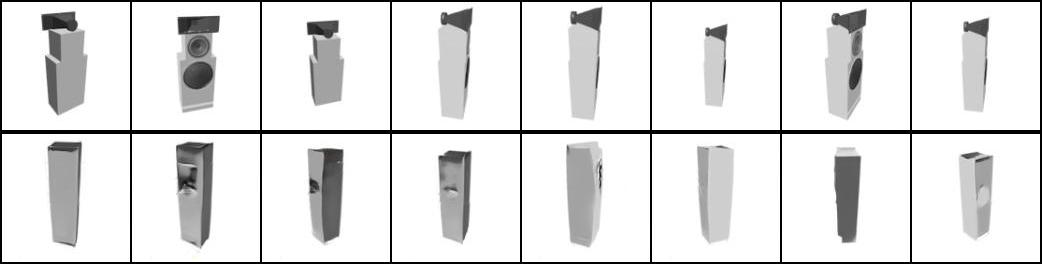}    \includegraphics[scale=0.28]{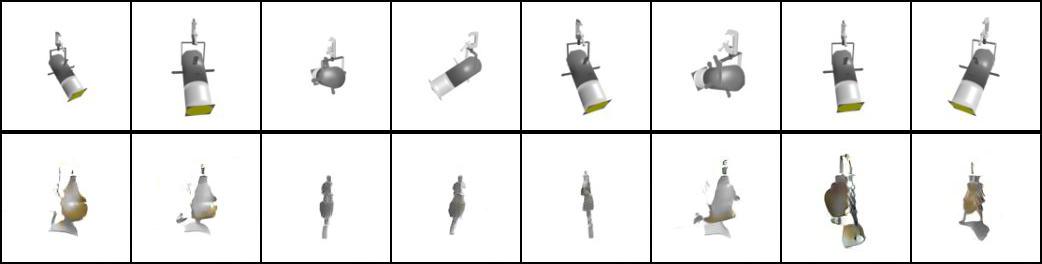}    \includegraphics[scale=0.28]{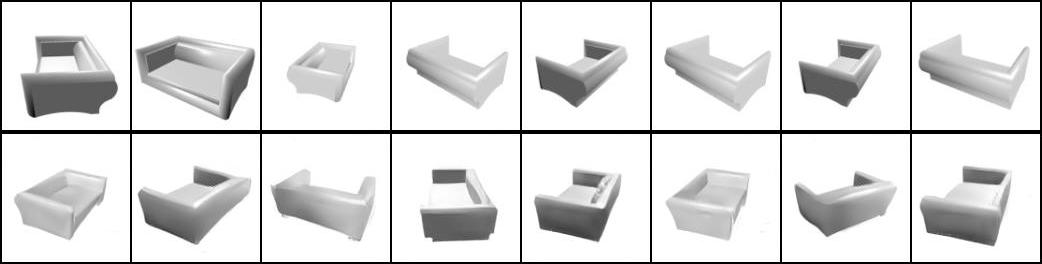}    \includegraphics[scale=0.28]{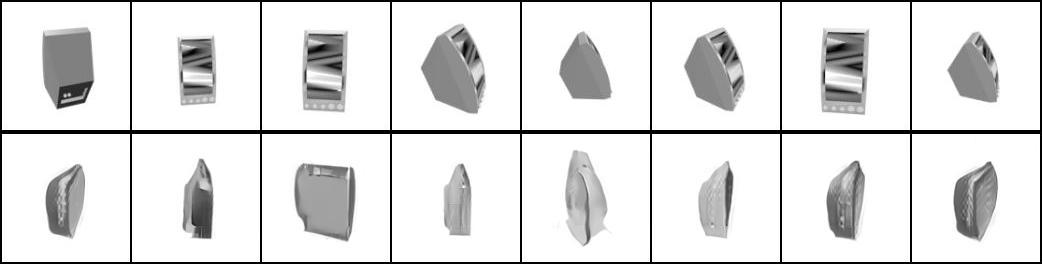}    \includegraphics[scale=0.28]{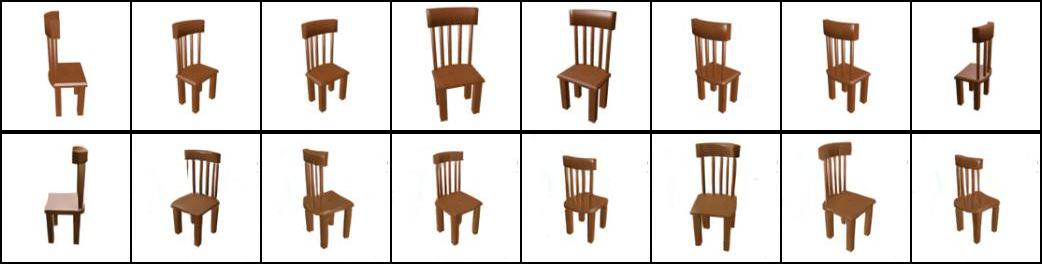}    
    \caption{More reconstructions of objects from the test split ShapeNet. For each block the top row is the input set and the bottom row is the reconstructed set.}
    \label{fig:shapenet_recons_more}
\end{figure}

\begin{figure}
    \centering
    \includegraphics[scale=0.28]{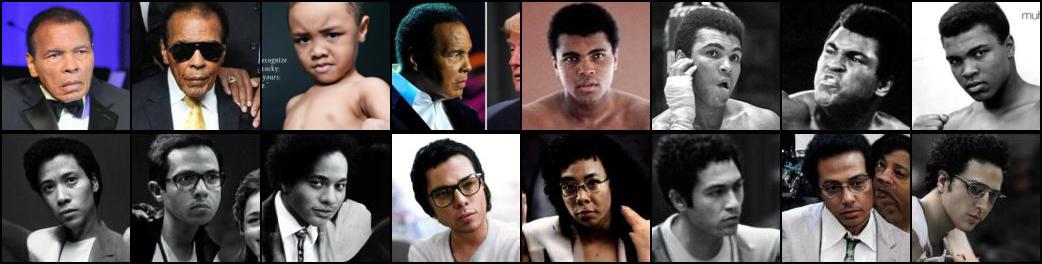}
        \includegraphics[scale=0.28]{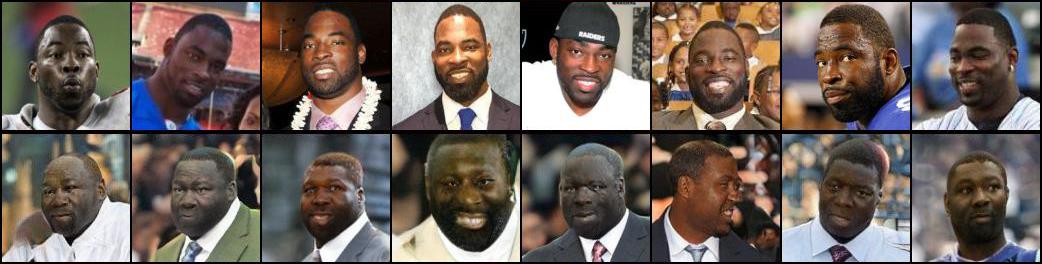}    \includegraphics[scale=0.28]{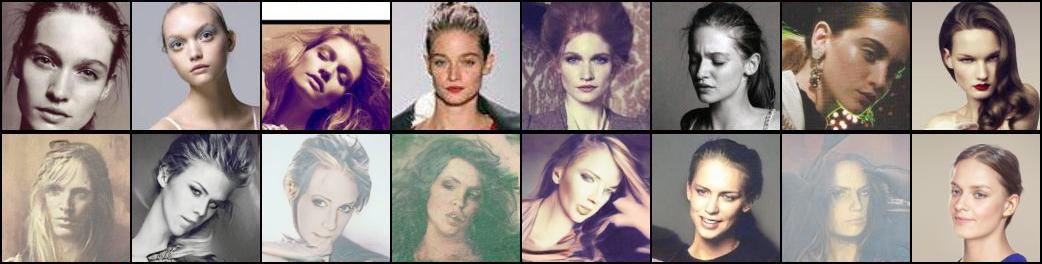}    \includegraphics[scale=0.28]{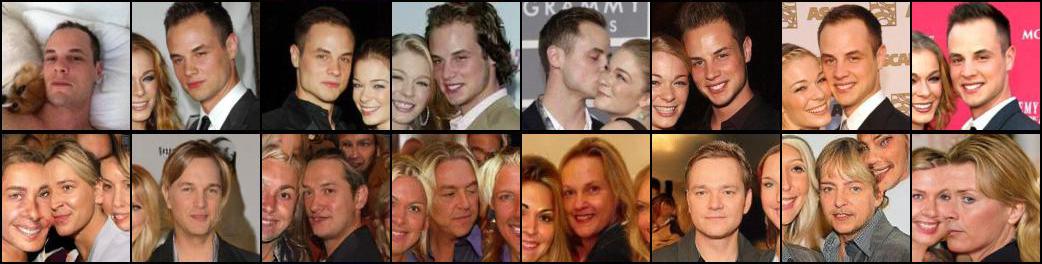}    \includegraphics[scale=0.28]{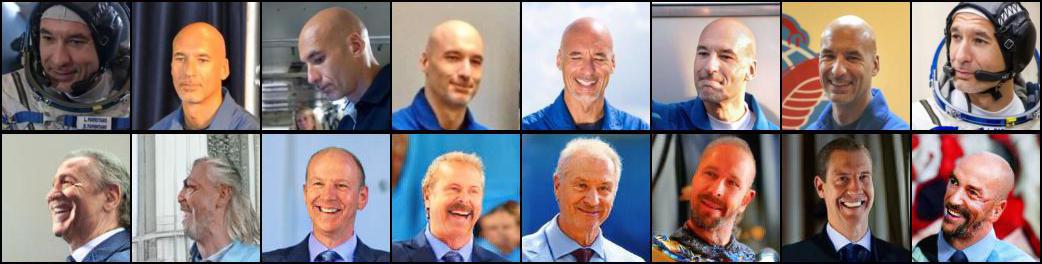}    \includegraphics[scale=0.28]{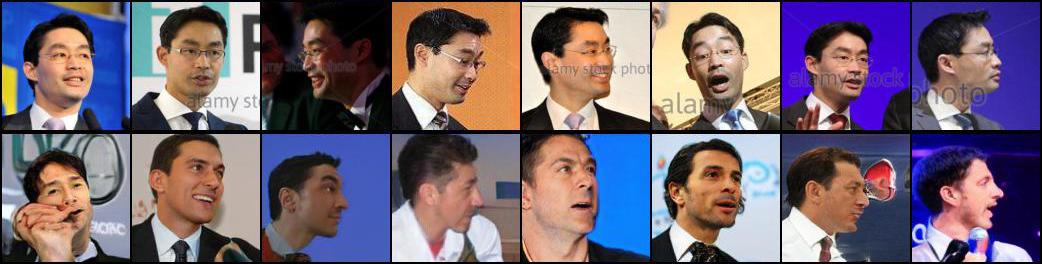}    \includegraphics[scale=0.28]{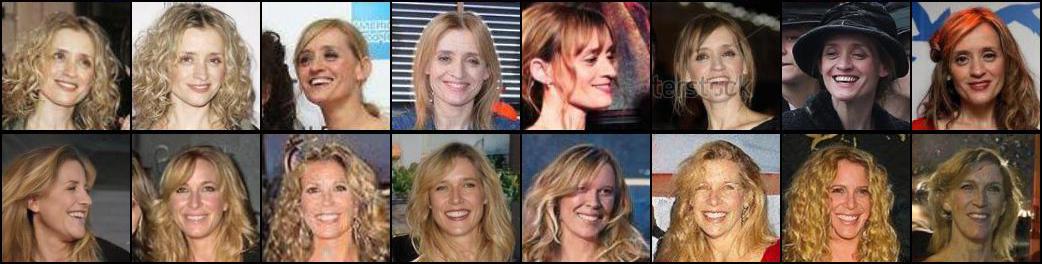}    \includegraphics[scale=0.28]{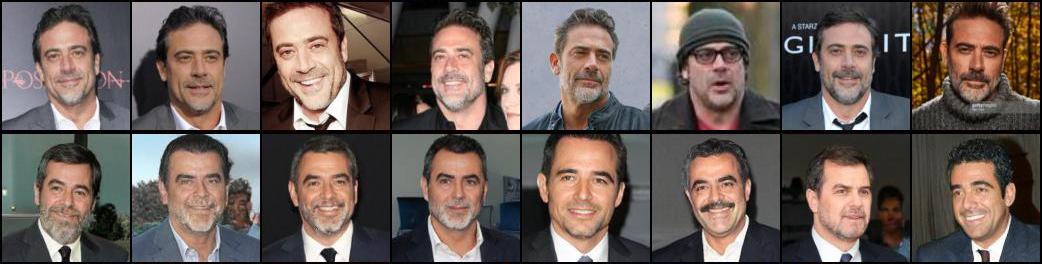}    
    \caption{More reconstructions of objects from the test split VggFace2. For each block the top row is the input set and the bottom row is the reconstructed set.}
    \label{fig:vggface_recons_more}
\end{figure}

\begin{figure}
    \centering
    \includegraphics[scale=0.18]{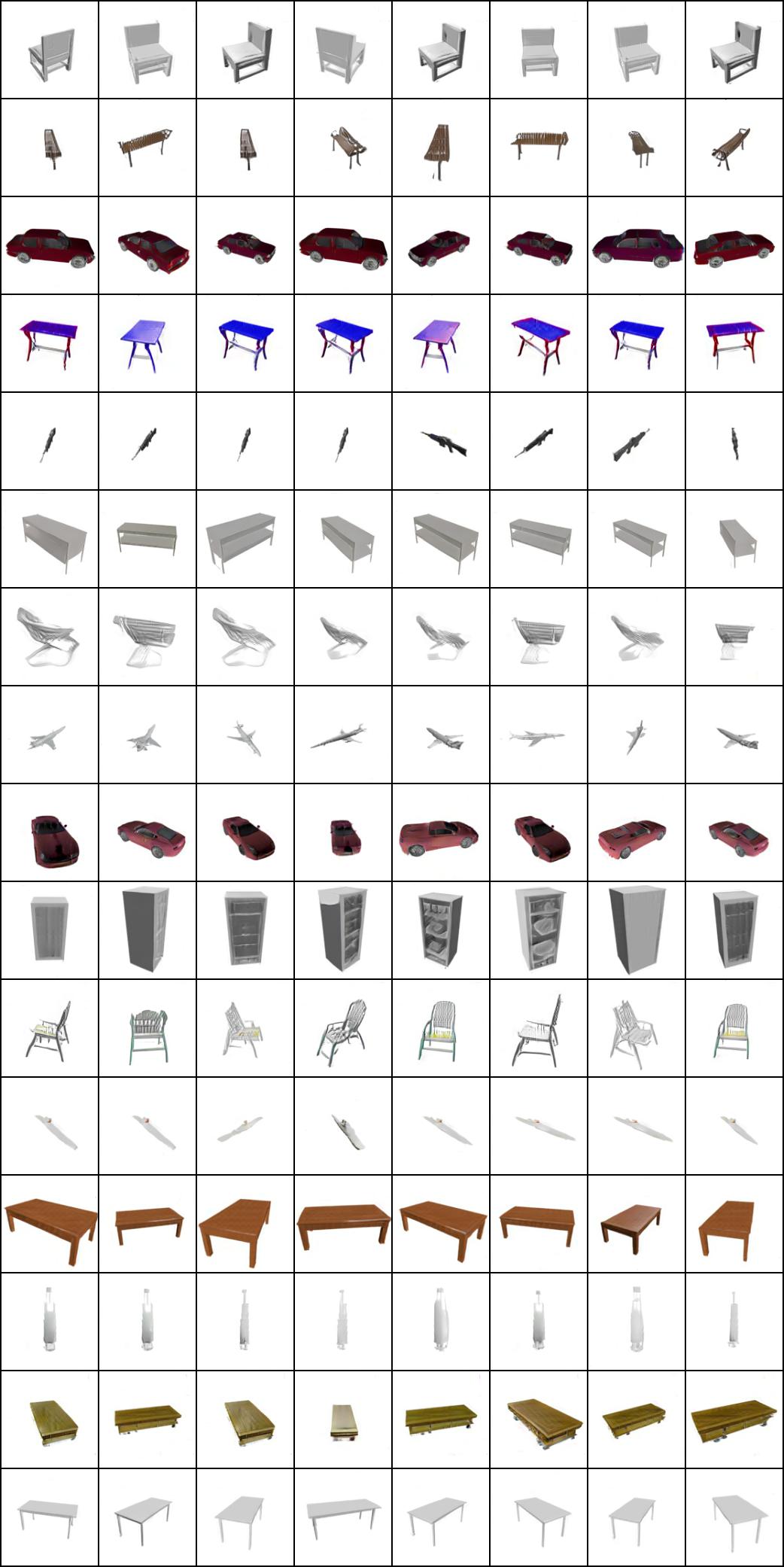}
        \includegraphics[scale=0.18]{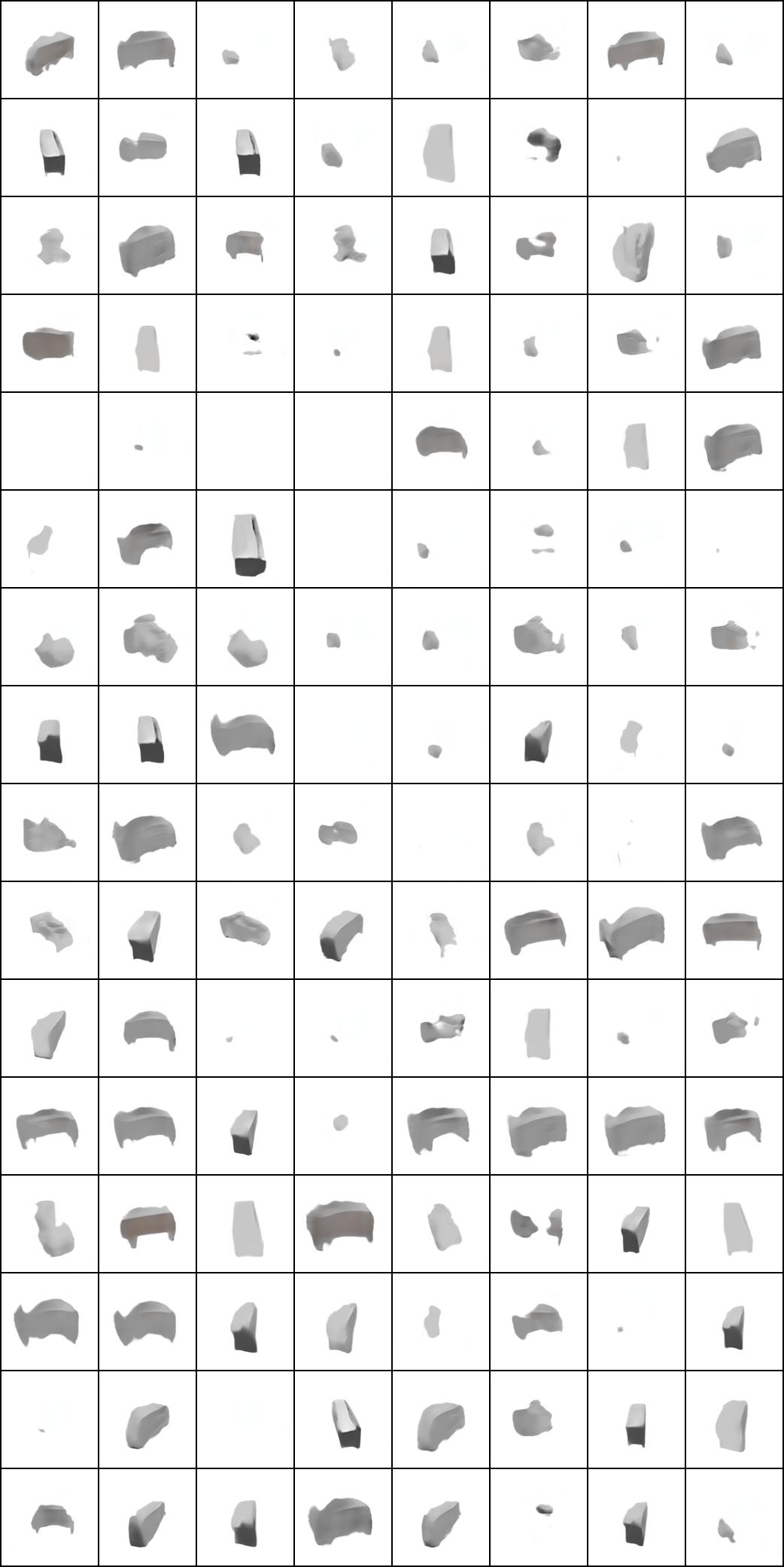}
    \caption{Uncurated ShapeNet samples from the learned autoregressive prior (left) and a uniform prior (right). }
    \label{fig:shapenet_free_sample_more}
\end{figure}

\begin{figure}
    \centering
    \includegraphics[scale=0.18]{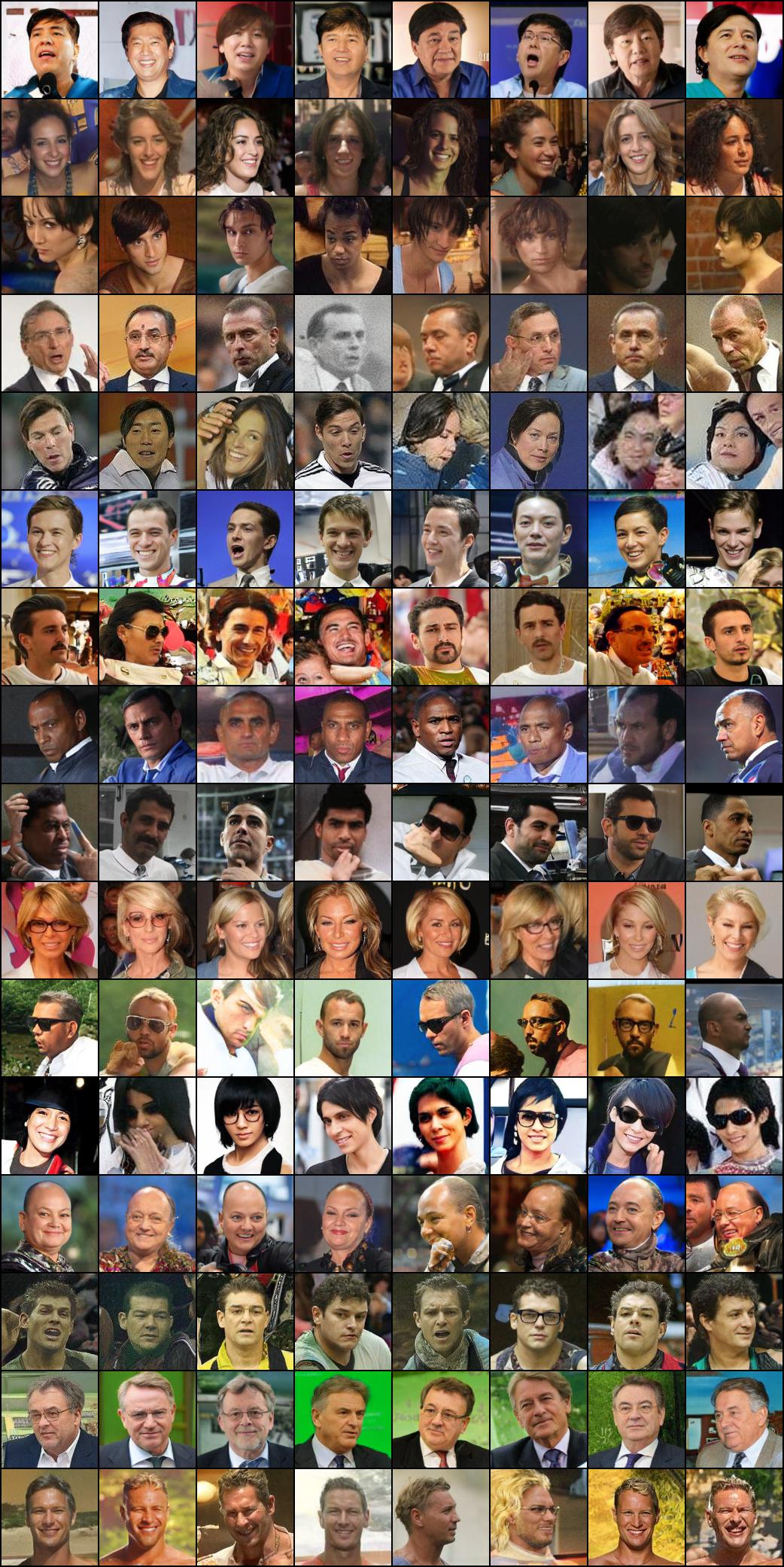}
        \includegraphics[scale=0.18]{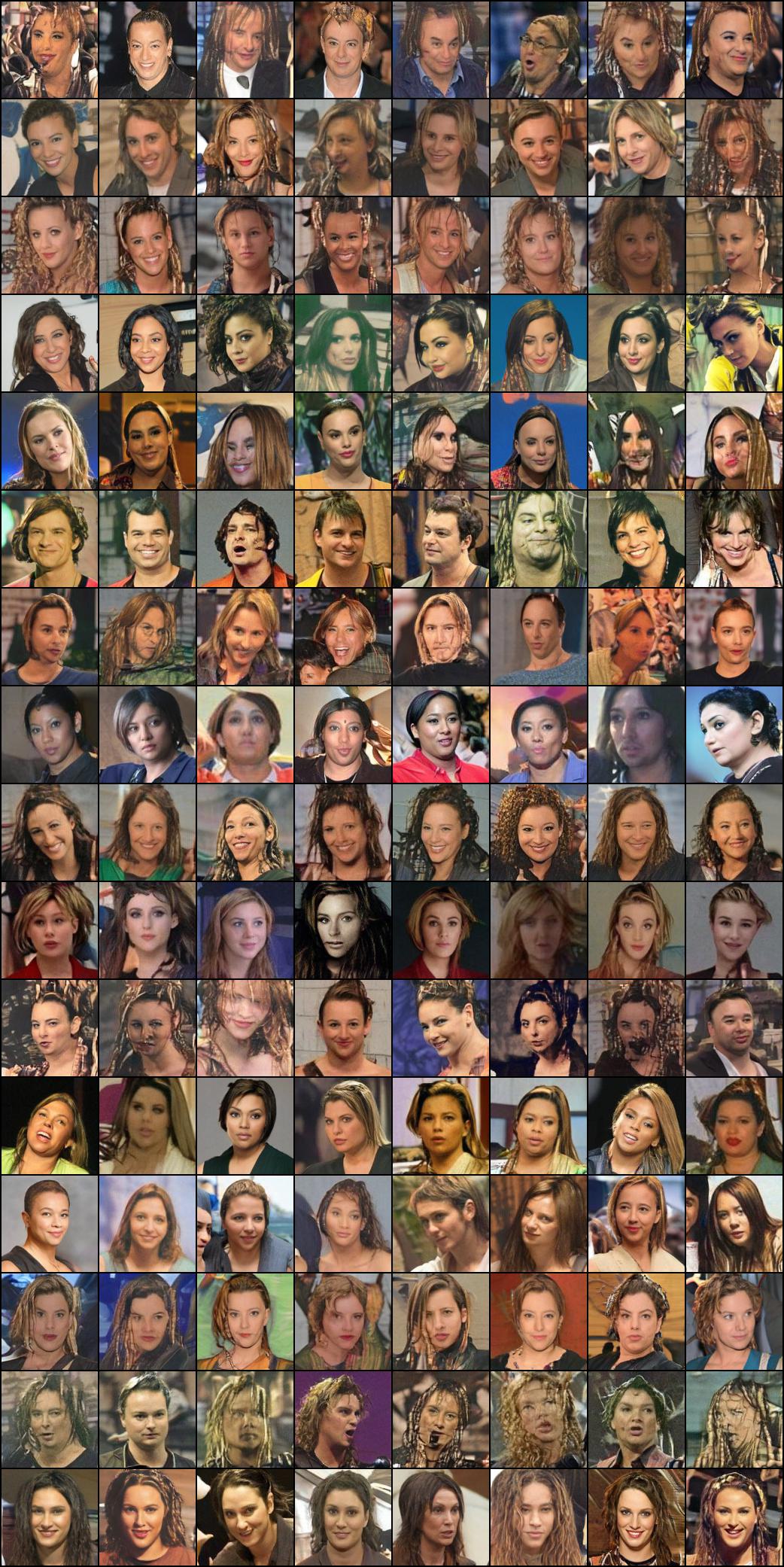}
    \caption{Uncurated VGGFace2 samples from the learned autoregressive prior (left) and a uniform prior (right). }
    \label{fig:vggface_free_sample_more}
\end{figure}

We first show more samples obtained with SDNs. Reconstructions of unseen sets can be found in Fig. \ref{fig:shapenet_recons_more} and Fig. \ref{fig:vggface_recons_more} for ShapeNet and VGGFace2, respectively. We also demonstrate that the learned prior produces good and  consistent samples in Figure \ref{fig:shapenet_free_sample_more} and Figure \ref{fig:vggface_free_sample_more}, where the uniform prior produces significantly worse samples. All the samples shown are uncurated.


\begin{figure}
    \centering
    \includegraphics[scale=0.28]{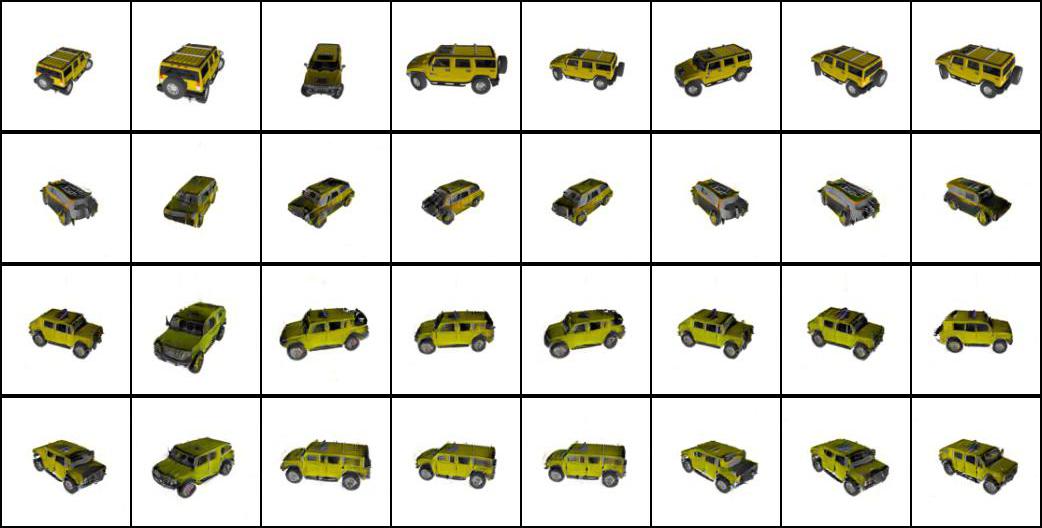}
    \includegraphics[scale=0.28]{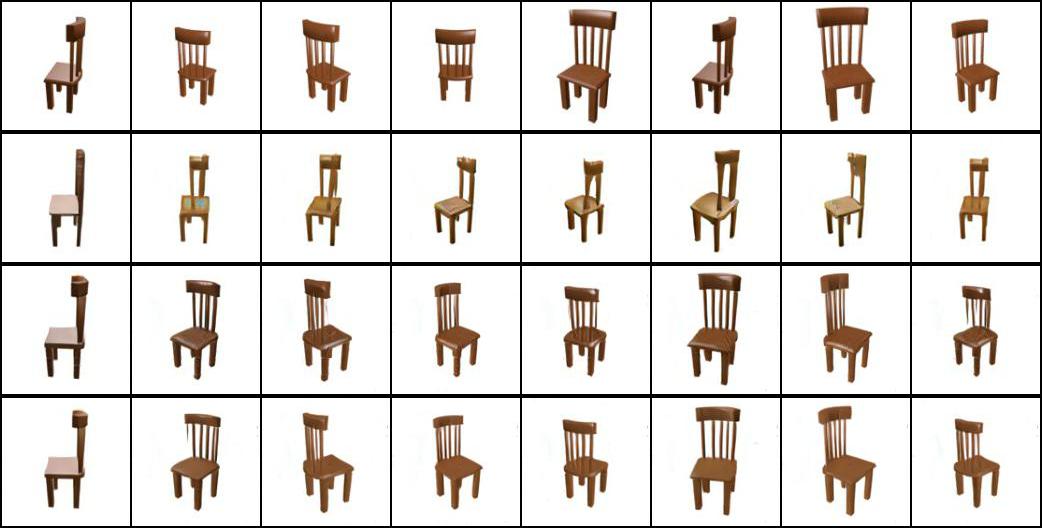}    \includegraphics[scale=0.28]{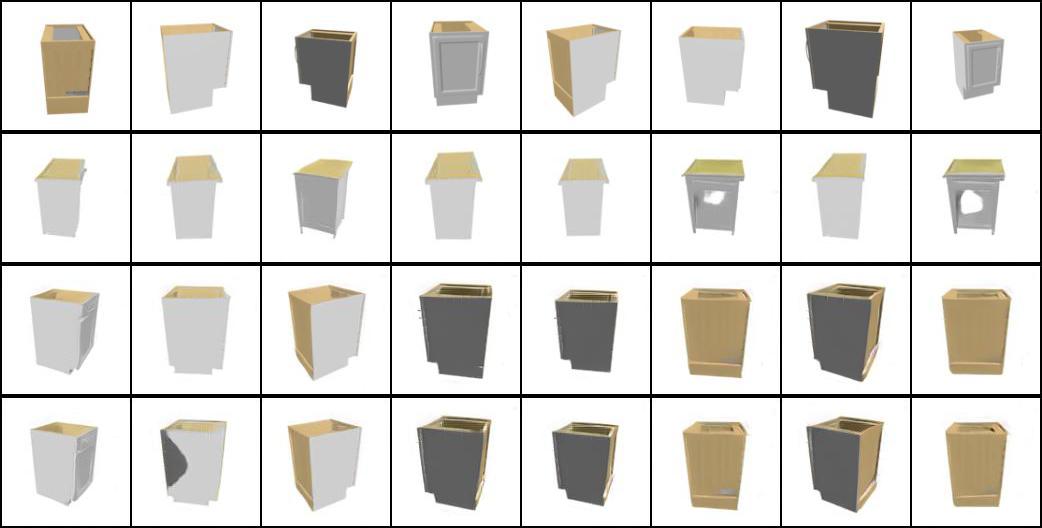}    \includegraphics[scale=0.28]{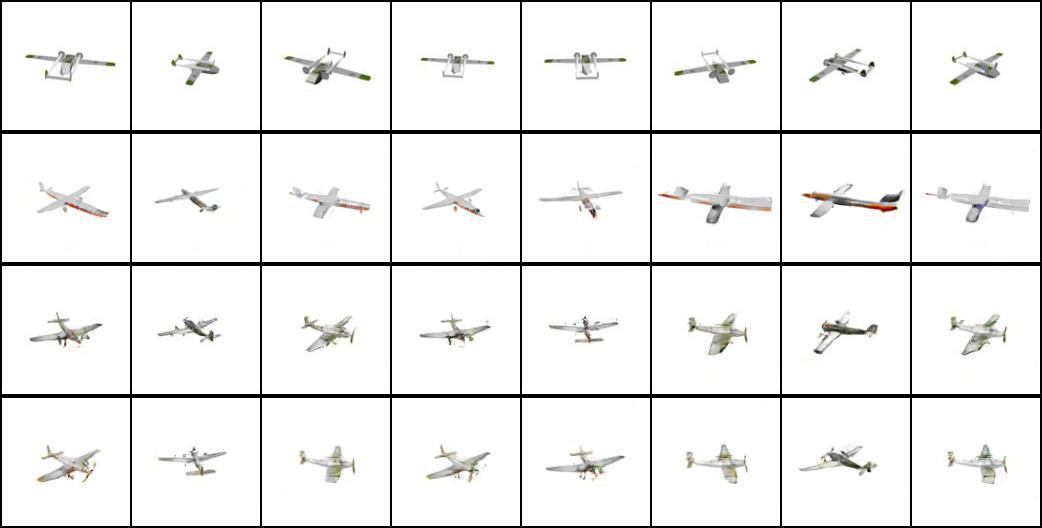}
   
    \caption{Reconstructions by varying input set size (see Table 1(b) in main text). For each block, the top row is the input set, with the following rows showing reconstructions obtained with the first, first 4 and all 8 views, respectively.}
    \label{fig:shapenet_setsize}
\end{figure}

\section{Varying input set size}
SDNs are trained with fixed set sizes (8 in our experiments). But because of the use of average pooling in the encoder, it is possible to test a trained SDN with varied input sizes (Table 1(b) in main text). We show four such results in Fig. \ref{fig:shapenet_setsize}. we see that an SDN is able to effectively utilize more input samples within the set and produce inreasingly more consistent reconstructions. 

\section{3D reconstruction on ShapeNet}
Our 3D reconstruction experiments show that SDN-reconstructed sets are accurate and consistent. In this section, we describe in more detail our extension of Occupancy Networks \cite{occnet} to deal with multiple input RGB views, which is not discussed in \cite{occnet}. Occupancy Networks use a convolutional encoder to compute a $d$-dimensional latent representation $\textbf{c} \in \mathbb{R}^d$ for a given RGB view.  This latent representation is then used as conditioning for an MLP $f_\theta: \mathbb{R}^3 \times \mathbb{R}^d \rightarrow \mathbb{R}$ that takes 3D points and predicts their occupancy (ie. whether the points lie inside or outside the object mesh).

Our hypothesis is that by average pooling the latent representations $\textbf{c}$ obtained by each element on a set (remember each element on a set corresponds to a different viewpoint of the same object) we can increase the amount of information about the object contained in $\textbf{c}$. However, that is only true if the latent representations $\textbf{c}$ of different views are in agreement, in other words, if they are different views of the \textit{same} object. To check this assumption we take a pre-trained Occupancy Network trained on single views (check \cite{occnet} for more details) and show in Tab. \ref{tab:occnetviews} that reconstruction accuracy as measured by IoU increases when the latent representations $\textbf{c}$ are pooled across views of the \textit{same} object\footnote{This evaluation was done on real sets of object views and following the evaluation protocol of \cite{occnet}.}. These results show that if elements of a SDN-reconstructed set are in agreement (eg. if a set contains different viewpoints of the same underlying object) reconstruction accuracy should improve when average pooling across elements in the set.

\begin{table*}[!h]
\centering 
\small
\begin{tabular}{|l|cc|}
\hline
 & \multicolumn{2}{|c|}{$\uparrow$\textbf{IoU}} \\ 
 & 1 view &     5 views   \\ \hline
Mean  &     0.593 &  0.621          \\ \hline

\end{tabular}
\caption{Aggregating multiple views for Occupancy Networks improves reconstruction accuracy.}
\label{tab:occnetviews}
\end{table*}

\begin{figure}[t]
    \centering
    \includegraphics[width=\textwidth]{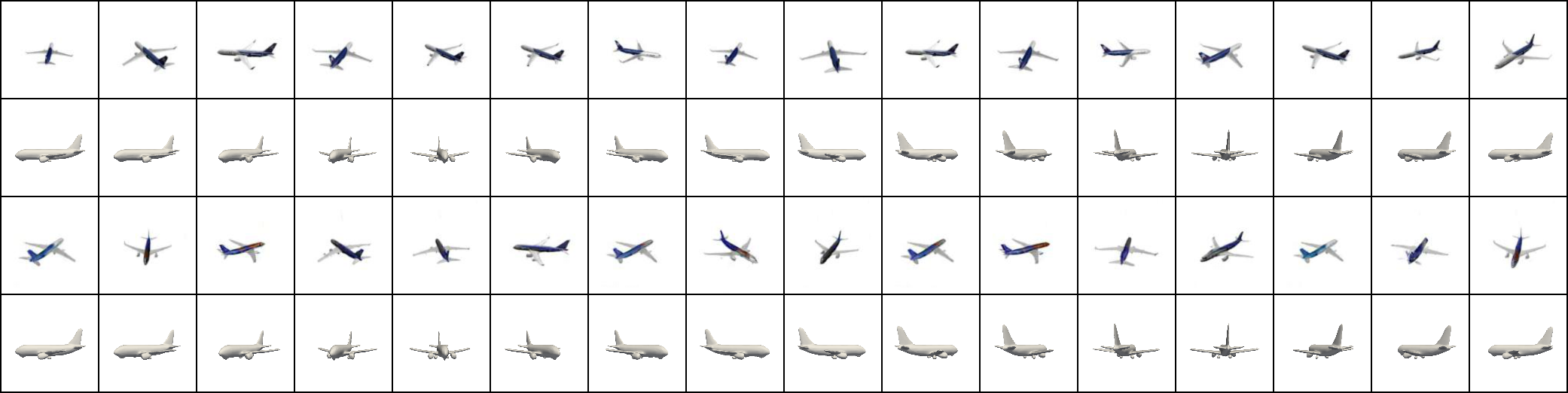} \\
    \includegraphics[width=\textwidth]{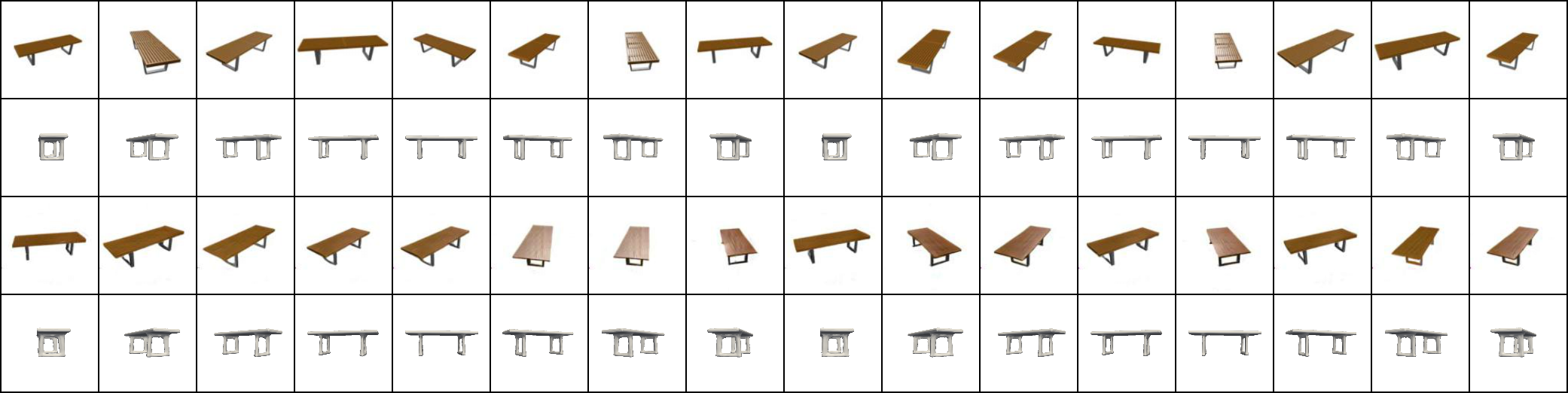} \\
    \includegraphics[width=\textwidth]{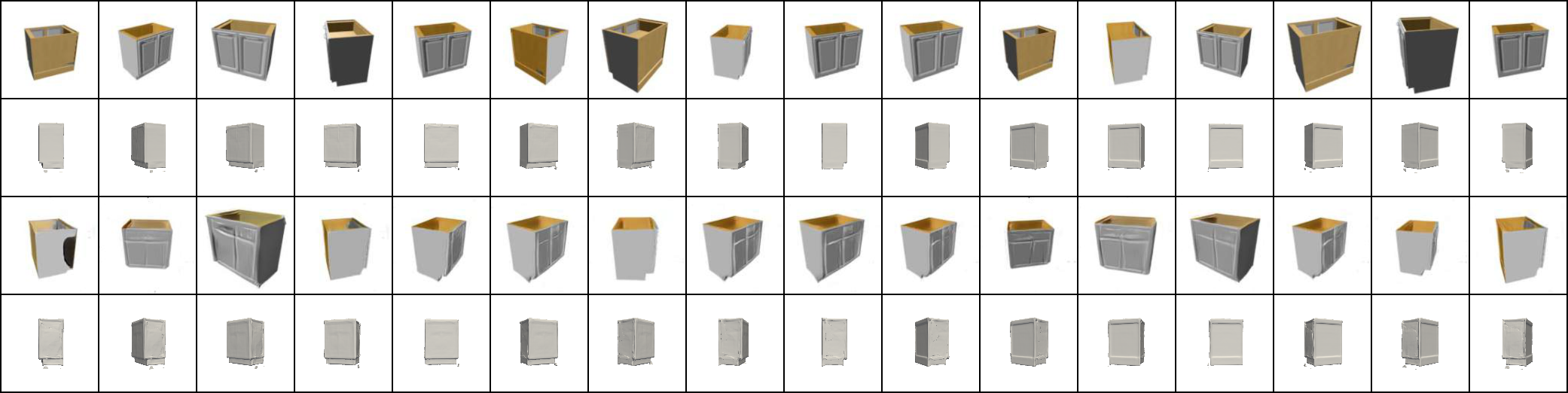} \\
    \includegraphics[width=\textwidth]{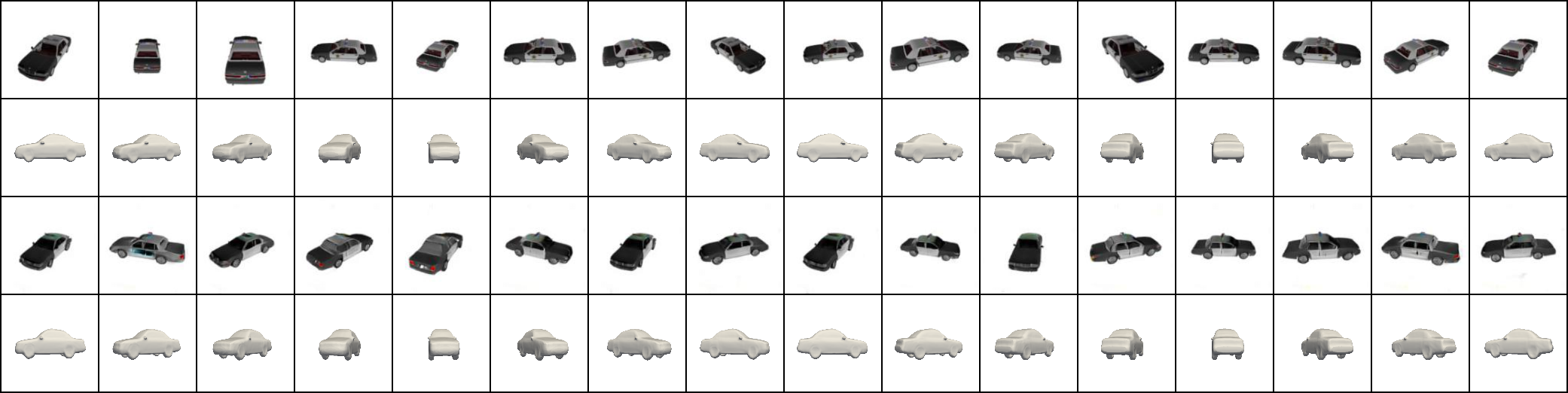} \\
    \includegraphics[width=\textwidth]{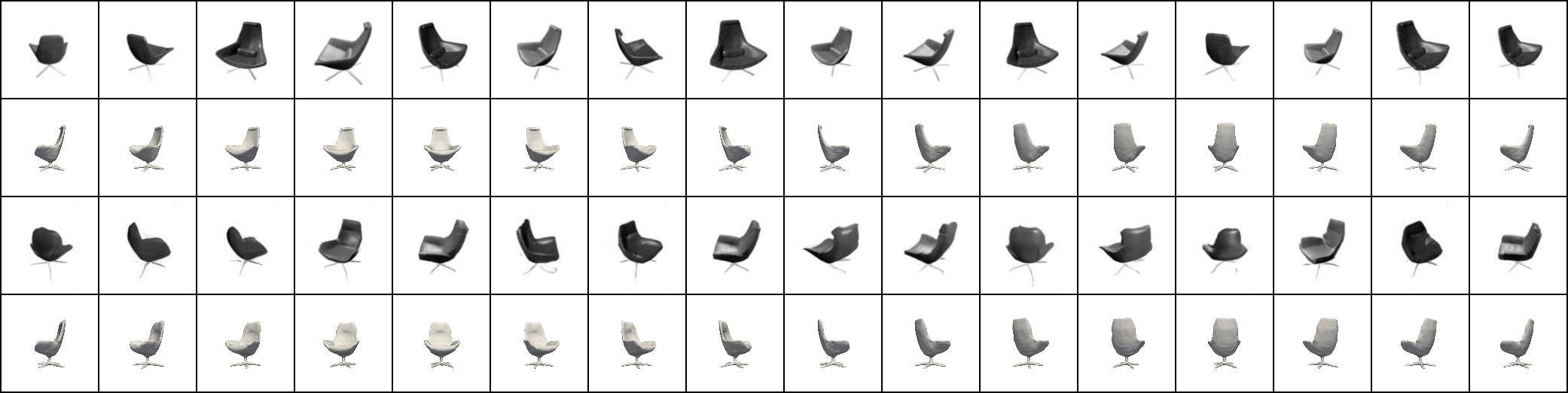} \\
    \includegraphics[width=\textwidth]{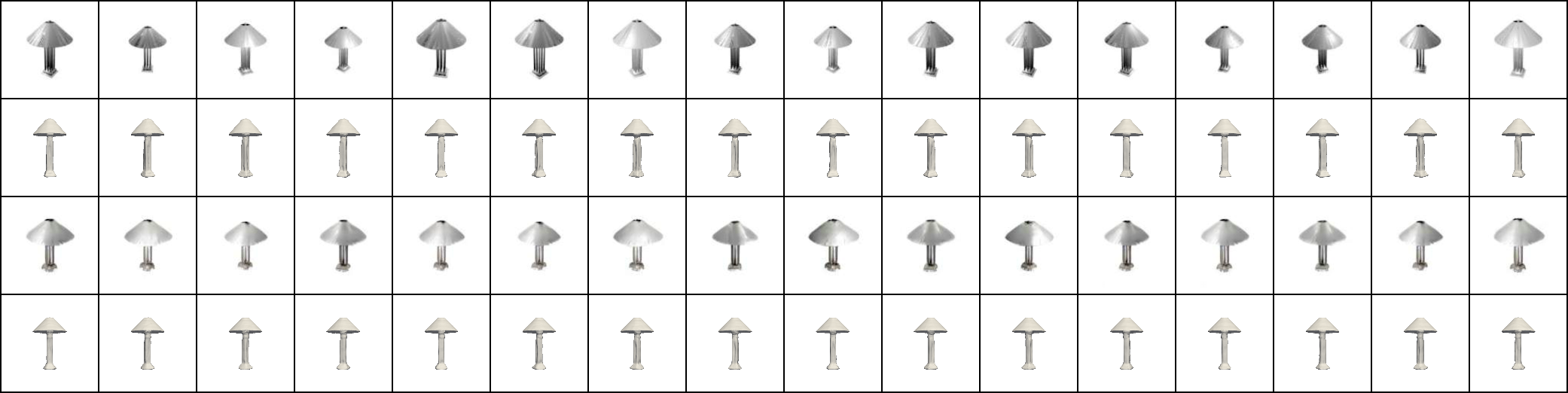} \\

    \caption{Random sampling of real and SDN-reconstructed sets with their corresponding 3D reconstructions obtained by our Occupancy Network extension. Each block shows an input views, input 3D meshes, reconstructed views, reconstructed 3D meshes, from top to bottom.}
    \label{fig:more_3d_reconstruction}
\end{figure}

In Fig. \ref{fig:more_3d_reconstruction} we show uncurated pairs of real and SDN-reconstructed sets together with their respective 3D reconstructions obtained by running our Occupancy Networks extension. For each grid the two first rows correspond to real set and an orbit of its predicted 3D reconstruction and the two last rows correspond to SDN-reconstructed set and its corresponding 3D reconstruction.

\begin{figure}[t]
    \centering
    \includegraphics[width=\textwidth]{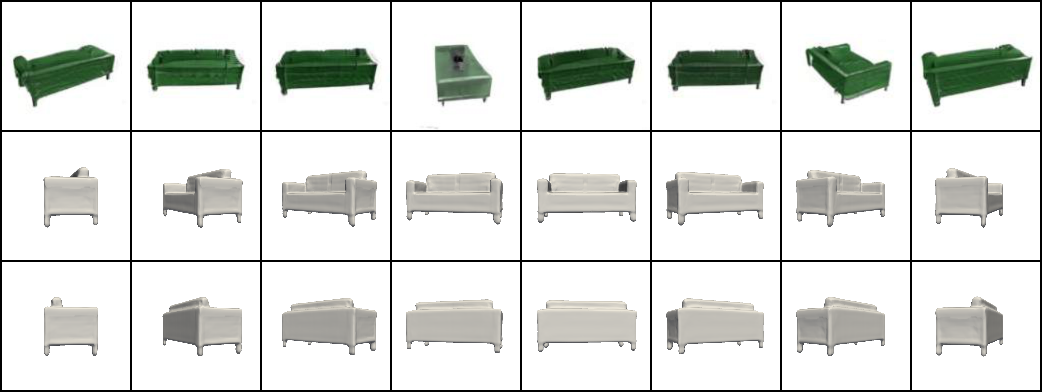} \\
    \includegraphics[width=\textwidth]{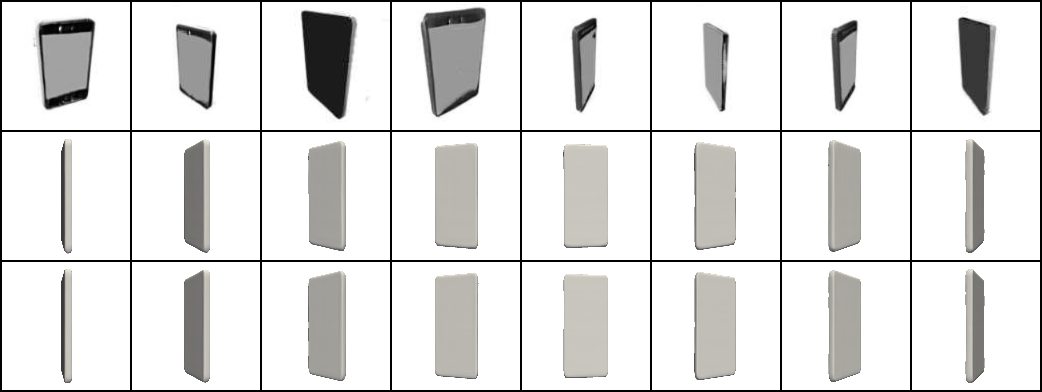} \\
    \includegraphics[width=\textwidth]{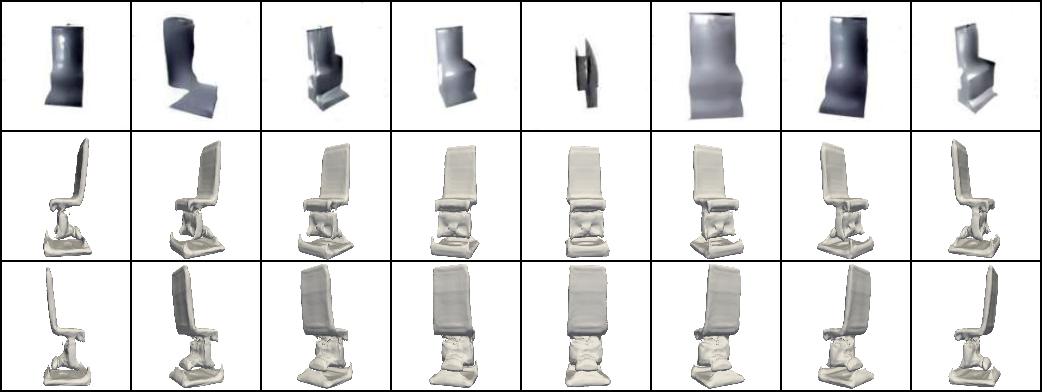} \\
    \includegraphics[width=\textwidth]{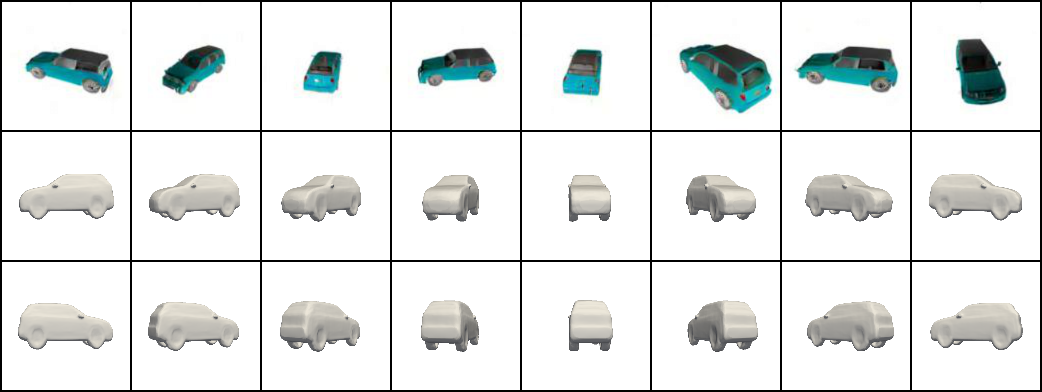}
    \caption{Sets sampled from the SDN prior with their corresponding 3D reconstruction. Each block shows a generated set consisting of 8 views and 16 of its reconstructed 3D mesh projections.}
    \label{fig:3d_reconstruction_free_samples}
\end{figure}

Finally, Fig. \ref{fig:3d_reconstruction_free_samples} shows sets sampled from the SDN prior and their corresponding 3D reconstruction. For each grid in the top row we show the sampled sets of 8 elements from the prior, and the bottom two rows show and orbit of the 3D mesh.

\section{Qualitative Analysis of Within Set Variation}

\begin{figure}[h]
\centering
\includegraphics[scale=0.22]{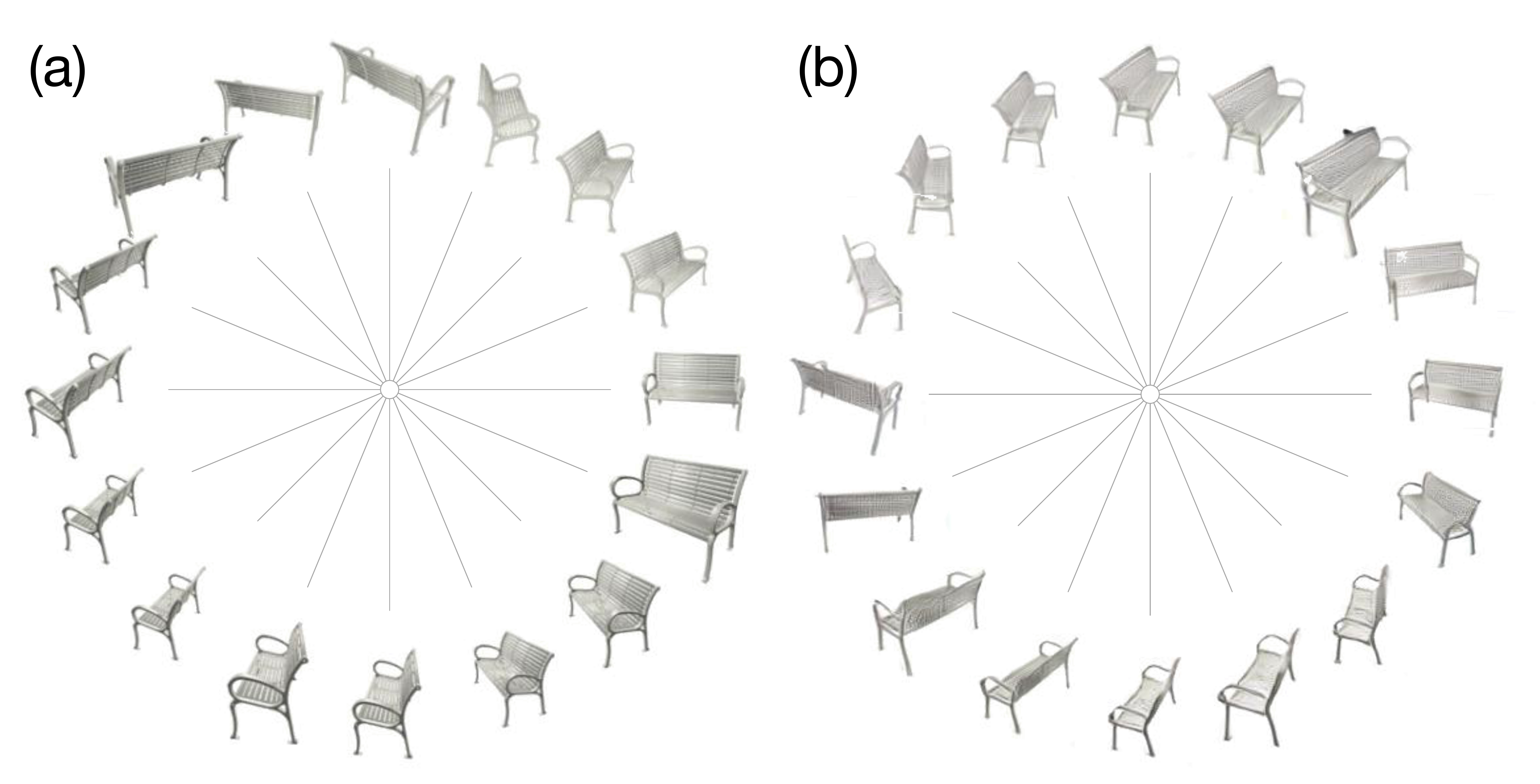}
\caption{Input (a) and reconstructed (b) sets from a ShapeNet bench where samples are arranged on a circle by manually estimated pose (see Supplementary Material for details). This visualization shows consistency in terms of appearance and pose variability between input and reconstructed sets.}
\label{fig:bench_wheel}
\end{figure}

In this section, we visualize the consistency and diversity of samples within a set produced by our SDN trained on ShapeNet object instances. To do so, for several object instances, we manually order samples by their pose (as estimated by a human annotator) to generate a continuum. We then map this continuum to a circle representing how the pose varies along the viewing sphere. By visualizing the input and reconstructed sets in this way we can examine how well the reconstructions capture the pose variability of the inputs. The reconstructions qualitatively capture the appearance of the input set (consistency) and the pose variation around the circle (diversity), with no sign of mode collapse. For the input set, we sampled 16 images randomly for that object. For the reconstructed set, we sampled 64 images, ordered them manually by visual inspection, and then decimated the continuum to obtain 16 reconstruction images.

\end{document}